\documentclass[sigconf]{acmart}
\AtBeginDocument{%
  }

\copyrightyear{2025}
\acmYear{2025}
\setcopyright{cc}
\setcctype{by}
\acmConference[ICVGIP 2025]{Indian Conference on Computer Vision, Graphics, and Image Processing}{December 17--20, 2025}{Mandi, India}
\acmBooktitle{Indian Conference on Computer Vision, Graphics, and Image Processing (ICVGIP 2025), December 17--20, 2025, Mandi, India}
\acmPrice{}
\acmDOI{10.1145/3774521.3774590}
\acmISBN{979-8-4007-1930-1/25/12}

\usepackage{cleveref}
\usepackage{subcaption}
\usepackage{siunitx}

\definecolor[named]{ForestGreen}{rgb}{0.13, 0.55, 0.13}
\newcommand{\B}[1]{\textbf{#1}}
\newcommand{\K}{{\Huge\(\blacksquare\)}}
\renewcommand{\k}{{\color{ForestGreen}\Huge\(\blacksquare\)}}
\renewcommand{\S}[1]{\hspace{#1\linewidth}}
\newcommand{\W}{{\Huge\(\square\)}}
\newcommand{\w}{{\color{ForestGreen}\Huge\(\square\)}}

\begin{document}

\title{Can Unsupervised Segmentation Reduce Annotation Costs for Video Semantic Segmentation?}

\author{Samik Some}
\email{samiks@iitk.ac.in}
\affiliation{
    \institution{IIT Kanpur}
    \city{Kanpur}
    \country{India}
}
\author{Vinay P. Namboodiri}
\email{vpn22@bath.ac.uk}
\affiliation{
    \institution{University of Bath}
    \city{Bath}
    \country{UK}
}

\renewcommand{\shortauthors}{S. Some \& V. P. Namboodiri}

\begin{abstract}
    Present-day deep neural networks for video semantic segmentation require a
    large number of fine-grained pixel-level annotations to achieve the best
    possible results. Obtaining such annotations, however, is very expensive. On
    the other hand, raw, unannotated video frames are practically free to obtain.
    Similarly, coarse annotations, which do not require precise boundaries, are
    also much cheaper. This paper investigates approaches to reduce the annotation
    cost required for video segmentation datasets by utilising such resources. We
    show that using state-of-the-art segmentation foundation models, Segment
    Anything Model (SAM) and Segment Anything Model 2 (SAM 2), we can utilise both
    unannotated frames as well as coarse annotations to alleviate the effort
    required for manual annotation of video segmentation datasets by automating
    mask generation. Our investigation suggests that if used appropriately, we can
    reduce the need for annotation by a third with similar performance for video
    semantic segmentation. More significantly, our analysis suggests that the
    variety of frames in the dataset is more important than the number of frames
    for obtaining the best performance.
\end{abstract}




\maketitle

\section{Introduction}
Semantic segmentation is a task where, given an image, we need to label each
pixel of the image as belonging to one of a given set of semantic classes.
Unlike instance segmentation, a key attribute of semantic segmentation is only
labelling each pixel as part of a semantic class and not differentiating
between objects within each class. While initially proposed for images, it was
only a short time before researchers looked into video semantic segmentation,
where instead of segmenting a single image, we have to segment all frames of a
given video clip. In order to train models for the aforementioned task, we need
access to large-scale, high-quality, manually annotated video datasets.

While there exist several such datasets, such as
Cityscapes~\citep{DBLP:conf/cvpr/CordtsORREBFRS16},
BDD100K~\citep{DBLP:conf/cvpr/YuCWXCLMD20},
CamVid~\citep{DBLP:conf/eccv/BrostowSFC08},
IDD~\citep{DBLP:conf/wacv/VarmaSNCJ19}, gathering such datasets for different
cities across the world in different weather and lighting conditions, and
manually annotating them in a fine-grained manner, is highly expensive.
Interestingly, the Cityscapes dataset also provides coarse annotations for a
much larger number of frames. These coarse annotations do not cover entire
objects and only roughly estimate their full masks. However, coarse annotations
are much cheaper to obtain. For Cityscapes, fine-grained annotations require an
average of 90 minutes per image, whereas coarse annotations only require 7
minutes per image to annotate~\citep{DBLP:conf/cvpr/CordtsORREBFRS16}. Apart
from fine-grained and coarse annotations, these datasets also provide us with a
very large number of unannotated, raw video frames.

Unfortunately, most deep neural network-based segmentation models only use the
finely annotated frames for training purposes. While there are some works that
attempt to utilise coarse annotations, they are quite few and are primarily
concerned with improving segmentation performance rather than data efficiency.
Recently, following the establishment of foundation models in natural language
processing tasks, such as large language models (LLMs), there has also been a
lot of research in establishing similar foundation models for vision tasks.
These include models such as CLIP~\citep{DBLP:conf/icml/RadfordKHRGASAM21},
SAM~\citep{DBLP:conf/iccv/KirillovMRMRGXW23}, SAM
2~\citep{DBLP:conf/iclr/RaviGHHR0KRRGMP25}, etc.. Of particular interest are
the Segment Anything Model (SAM) and its successor, Segment Anything Model 2
(SAM 2), which are foundation models specialising in image and video
segmentation, respectively. Extensively trained on huge datasets, these models
are extremely performant and can be trained on various downstream segmentation
tasks. However, they are also very large. Even fine-tuning said models from end
to end requires a lot of computational resources, which may not always be
available. Running them in inference mode, however, is quite cheap
comparatively.

In this paper, we consider using these models to alleviate the annotation
efforts required for video semantic segmentation datasets. Essentially, our
goal is to find ways to use these foundation models to generate pseudo-labels
from unannotated frames or refine the coarse annotations already available,
such that using this augmented data helps us achieve similar performance to
using completely manually annotated data. We aim to reduce the number of
manually annotated samples rather than improve the segmentation performance. To
this end, we perform a variety of experiments to investigate the appropriate
use of the Segment Anything models and present the ways we found that help us
achieve our goal. We also report on several other approaches we tried that did
not produce good results. The detailed analysis of all these experiments is
provided and discussed in \cref{sec:experiments}.

Through this work, we make the following contributions:
\begin{itemize}
    \item We show that the variety of densely annotated video frames is more important
          than the number of densely annotated video frames through a carefully
          constructed set of experiments.
    \item We find that auto-annotation using Segment Anything models is applicable for
          unannotated frames as well as coarse annotations.
    \item We evaluate our method on the Cityscapes and IDD datasets using multiple
          semantic segmentation networks. Note that, as our primary focus is on reducing
          the annotation cost, we do not evaluate \textit{all} the various methods that
          exist in the literature.
\end{itemize}

\section{Related Works}
\subsection{Image Semantic Segmentation}
Semantic segmentation as a problem was first defined on images, where the goal
is to assign every pixel of an image a class label from a pre-defined list of
semantic classes. The community has researched semantic segmentation in images
for a long time, and there are several surveys, such as
\citet{DBLP:journals/ijmir/GuoLGL18, DBLP:journals/ijon/HaoZG20}, which
document the various techniques used for the same. Some relatively newer
deep-learning based techniques include \citet{DBLP:journals/corr/ChenPKMY14,
    DBLP:conf/cvpr/LongSD15, DBLP:conf/cvpr/YuKF17, DBLP:conf/cvpr/ZhaoSQWJ17,
    DBLP:journals/pami/ChenPKMY18, DBLP:conf/cvpr/YangYZLY18}. These works
primarily utilise Convolution Neural Network (CNN) backbones such as
VGG~\citep{DBLP:journals/corr/SimonyanZ14a},
ResNet~\citep{DBLP:conf/cvpr/HeZRS16}, ResNeXt~\citep{DBLP:conf/cvpr/XieGDTH17}
to extract features, which are then classified into the required classes to
generate a low-resolution segmentation mask and upsampled back to the original
resolution. The papers improve their performances primarily via better ways of
extracting features and combining them before classification with techniques
such as dilated convolutions, pyramid pooling, atrous spatial pyramids,
conditional random fields, etc.

Some more recent approaches, such as \citet{DBLP:conf/nips/ChengSK21,
    DBLP:conf/nips/XieWYAAL21, DBLP:conf/iccv/StrudelPLS21,
    DBLP:conf/cvpr/GuKW00CLCP22, DBLP:conf/iccvw/JainSOHL0S23,
    DBLP:conf/cvpr/Jain0C0OS23} are based on Transformer networks, introduced by
\citet{DBLP:conf/nips/VaswaniSPUJGKP17}, instead of CNNs. In general, these
models perform much better than earlier works. These approaches utilise
pipelines where image features are extracted using either a transformer encoder
or CNN, and then classified using a transformer decoder or full
transformer-based pipelines with
ViT-like~\citep{DBLP:conf/iclr/DosovitskiyB0WZ21} architectures. Several of
these approaches also move away from directly generating the segmentation mask
and instead opt for mask classification, where several binary masks are
generated and then combined to produce the final segmentation output.

\subsection{Video Semantic Segmentation}
Semantic segmentation has also been applied to video clips for quite some time
with works such as \citet{DBLP:conf/iccv/GaddeJG17, DBLP:conf/cvpr/LiuWQYBS17,
    DBLP:conf/cvpr/NilssonS18, DBLP:conf/cvpr/JainWG19, DBLP:conf/icmcs/LeeCP21,
    DBLP:conf/eccv/LiuSYW20}, focusing primarily on using optical flow features for
capturing temporal relationships between frames. Generally, these flow features
are used to propagate the segmentation masks from one frame to the next.
However, optical flows are often expensive to compute and do not predict well
beyond a few frames.

Newer approaches such as, \citet{DBLP:conf/cvpr/LiSL18,
    DBLP:conf/cvpr/HuCWLSP20, DBLP:conf/cvpr/SunLDPG22, DBLP:conf/icip/WangW021},
do not use optical flow features. Instead, these approaches directly learn
temporal dependencies using their networks. \citet{DBLP:conf/cvpr/LiSL18}
utilise a combination of low-cost, low-level and high-cost, high-level feature
extractors and feature propagation to selectively compute high-cost, high-level
features only for certain key-frames. In other cases, the high-level features
are propagated from earlier frames using spatially variant convolutions.
\citet{DBLP:conf/cvpr/HuCWLSP20} spread out feature extraction over multiple
frames to reduce computation costs and aggregate them for mask prediction.
\citet{DBLP:conf/cvpr/SunLDPG22} extract coarse to fine features from earlier
to more recent frames and then use a cross-frame feature mining module to
improve the target frame features before predicting the segmentation mask.
\citet{DBLP:conf/icip/WangW021} extract features from multiple past frames and
use a temporal memory attention module to model temporal relationships.

\subsection{Unannotated Frames and Coarse Annotations}
All the works discussed in the earlier sections utilise only fine-grained
annotations for training purposes. Unfortunately, obtaining fine-grained human
annotations, especially for video clips with 20-60 frames per second, is very
expensive. This is the primary reason most video segmentation datasets, such as
Cityscapes~\citep{DBLP:conf/cvpr/CordtsORREBFRS16},
BDD100K~\citep{DBLP:conf/cvpr/YuCWXCLMD20},
CamVid~\citep{DBLP:conf/eccv/BrostowSFC08},
IDD~\citep{DBLP:conf/wacv/VarmaSNCJ19}, only have fine-grained labels for
specific frames. While Cityscapes provides coarse annotations for a much larger
number of frames, the other datasets do not. However, all of them provide a
large number of unannotated frames.

Semi-supervised segmentation techniques attempt to utilise unannotated images
to generate pseudo-labels for further training. Works in this space include
\citet{DBLP:conf/iccv/SoulySS17, DBLP:conf/bmvc/HungTLL018,
    DBLP:journals/pami/MittalTB21, DBLP:conf/eccv/ChenLCCCZAS20,
    DBLP:conf/cvpr/0001SRSNTC19, DBLP:journals/corr/abs-2004-14960,
    DBLP:conf/eccv/LuoY20, DBLP:conf/iclr/ZouZZLBHP21,
    DBLP:conf/cvpr/WangWSFLJWZL22}. These works involve using GANs to generate
additional training data~\citep{DBLP:conf/iccv/SoulySS17}, using discriminators
to differentiate between predicted and ground-truth
masks~\citep{DBLP:conf/bmvc/HungTLL018, DBLP:journals/pami/MittalTB21} as well
as iterative self-training~\citep{DBLP:conf/cvpr/0001SRSNTC19,
    DBLP:conf/eccv/ChenLCCCZAS20, DBLP:journals/corr/abs-2004-14960} following a
teacher student strategy. \citet{DBLP:conf/iclr/ZouZZLBHP21} proposes using
decoder prediction and Grad-CAM along with a calibrated fusion module to
produce pseudo-labels. While most of these methods rely on some confidence
level to filter out ambiguous pseudo-labels,
\citet{DBLP:conf/cvpr/WangWSFLJWZL22} go one step further and suggest training
with said unreliable pixels. Their strategy is based on the observation that
even unreliable labels can be used as negative samples for several classes.

Other works, such as \citet{DBLP:conf/wacv/DasXHAS23}, aim to use only coarse
annotations to reach similar performance levels as fine-grained annotations in
image semantic segmentation tasks. In order to do this, they utilise synthetic
video segmentation datasets, which can provide dense labels at negligible cost,
and self-training to iteratively improve their predictions for unknown regions
in the coarse annotations.

\subsection{Vision Foundation Models}
Recently, after the explosion of large Transformer-based models in natural
language processing, there has been progressive research in establishing large,
performant models with extensive pre-training, which work well on several
downstream tasks, known as foundation models. In the realm of vision, these
include models such as CLIP~\citep{DBLP:conf/icml/RadfordKHRGASAM21}, which is
a vision language model pre-trained with contrastive learning between
image-caption pairs.

For semantic segmentation, at least two papers are considered foundation
models. These are Segment Anything Model
(SAM)~\citep{DBLP:conf/iccv/KirillovMRMRGXW23} and Segment Anything Model 2
(SAM 2)~\citep{DBLP:conf/iclr/RaviGHHR0KRRGMP25}. While SAM is limited to
images, SAM 2 can segment videos as well. Both of these models can take a wide
variety of inputs, including bounding boxes, points, masks, and theoretically
even text prompts. They are pre-trained on large datasets and produce excellent
segmentation results.

\begin{figure}[t]
    \centering
    \begin{subfigure}{0.495\linewidth}
        \includegraphics[width=\linewidth]{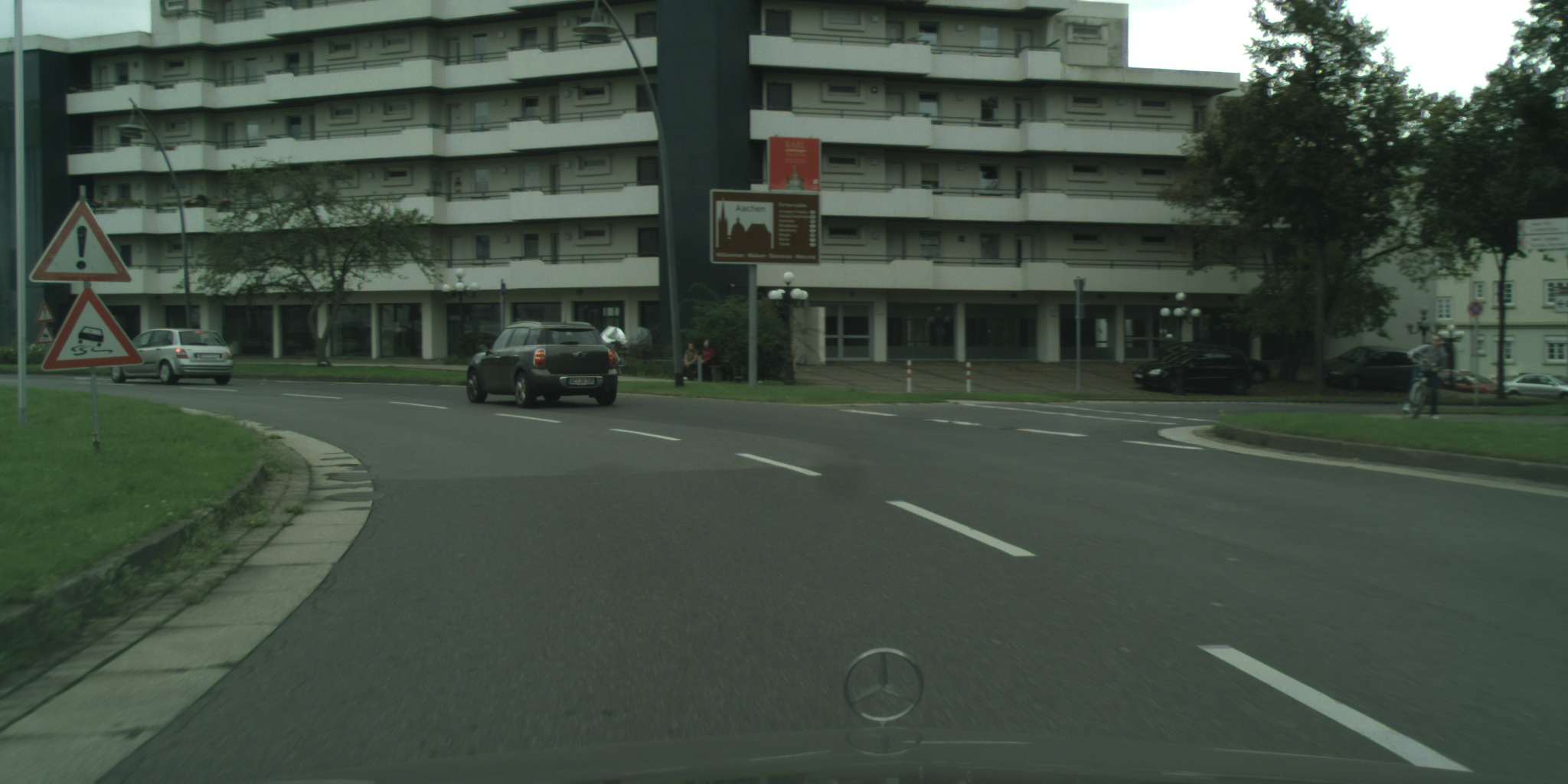}
        \caption{Image}
    \end{subfigure}
    \begin{subfigure}{0.495\linewidth}
        \includegraphics[width=\linewidth]{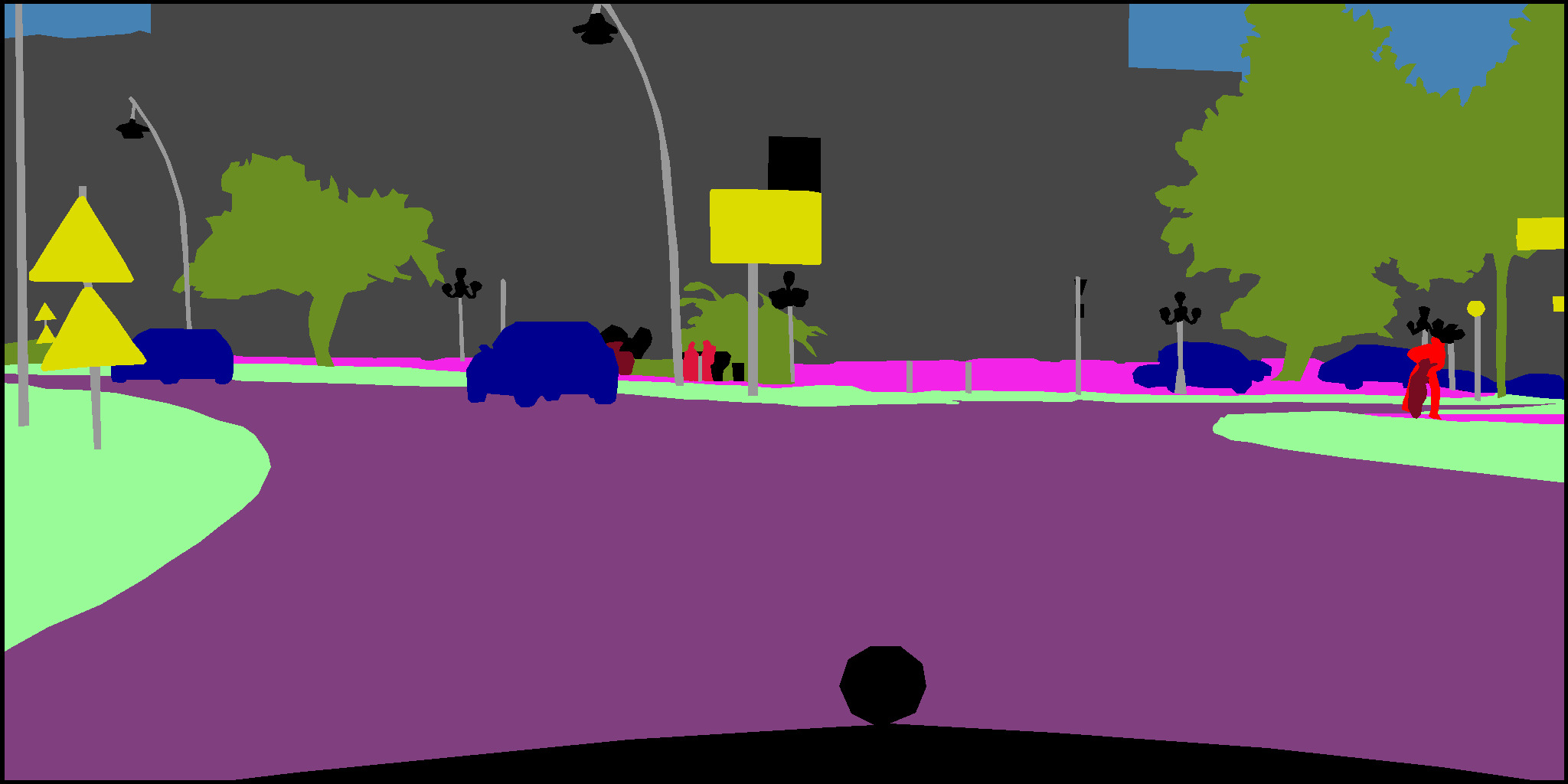}
        \caption{Fine-grained annotation}
    \end{subfigure}

    \begin{subfigure}{0.495\linewidth}
        \includegraphics[width=\linewidth]{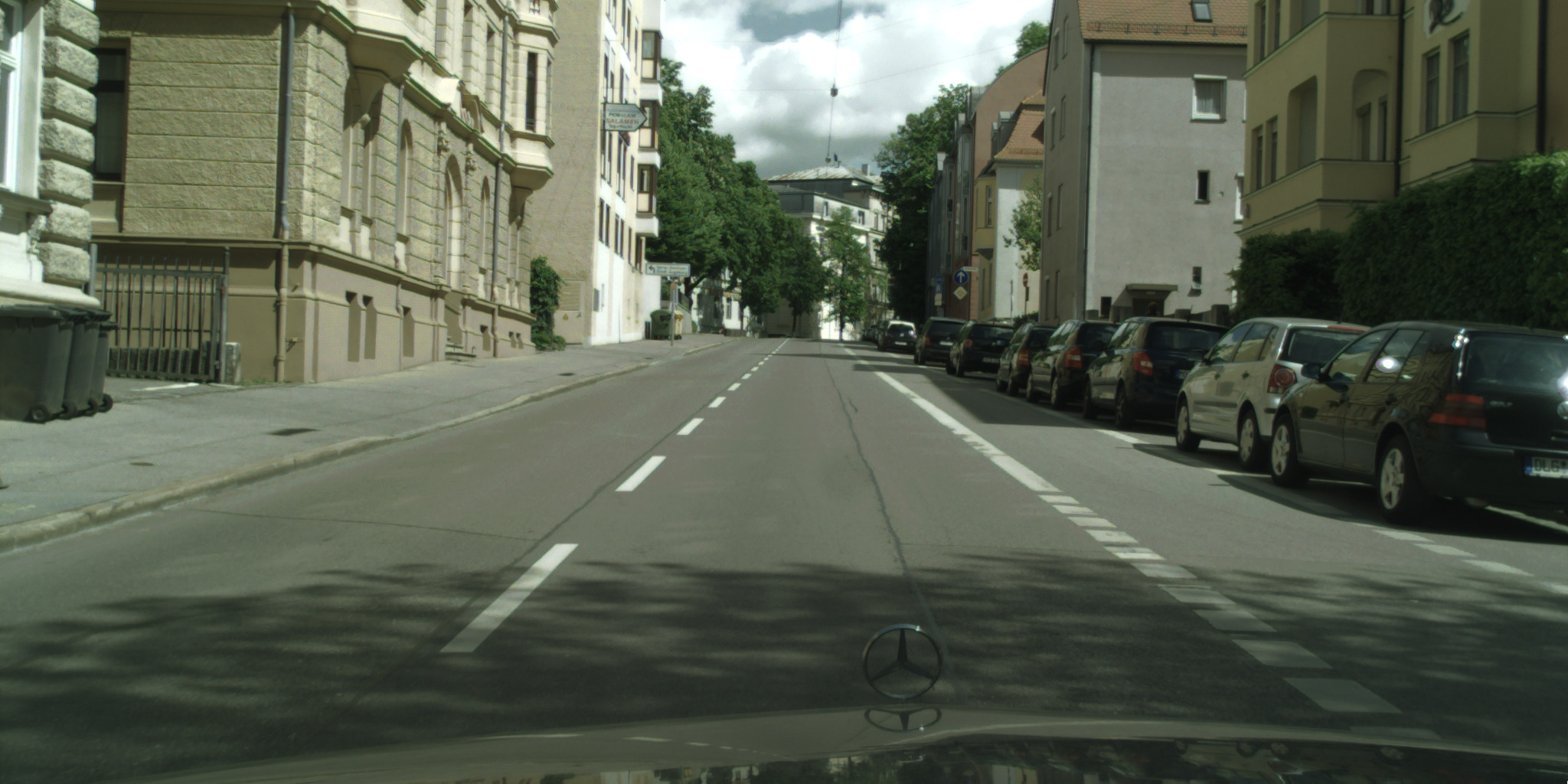}
        \caption{Image}
    \end{subfigure}
    \begin{subfigure}{0.495\linewidth}
        \includegraphics[width=\linewidth]{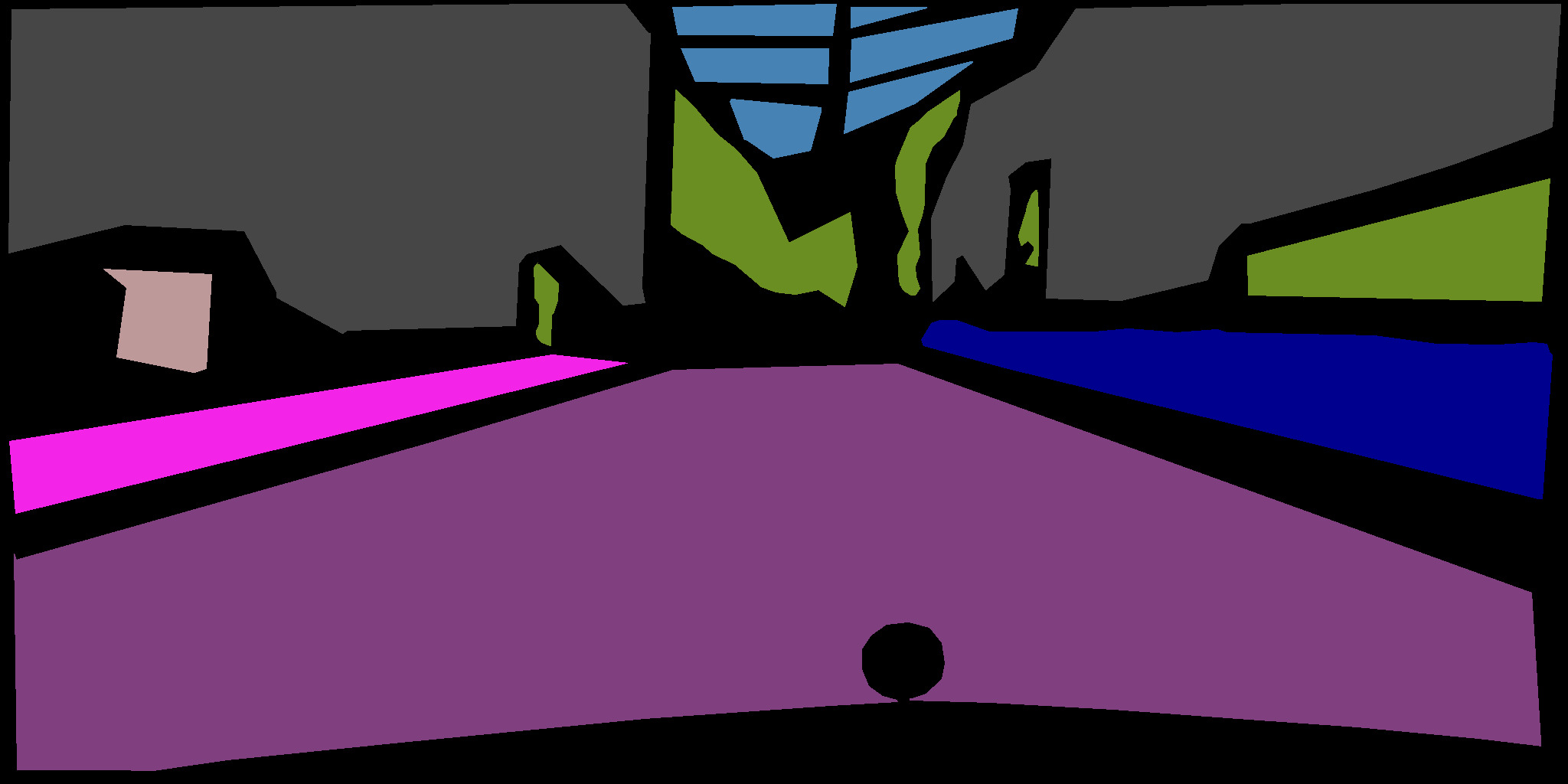}
        \caption{Coarse annotation}
    \end{subfigure}
    \caption{Examples of fine-grained and coarse annotations in the Cityscapes dataset.}
    \Description{Examples of fine-grained and coarse annotations in the Cityscapes
        dataset.} \label{fig:cityscapes_annotation_example}
\end{figure}

\section{Dataset and Model}
As we experiment with both unannotated and coarsely annotated frames, we wanted
to use a dataset that provides both. Unfortunately, while several video
segmentation datasets exist, only
Cityscapes~\citep{DBLP:conf/cvpr/CordtsORREBFRS16} provides coarse annotations.
The dataset is comprised of video clips from a dashboard-mounted camera inside
a vehicle driven in several German cities. It provides 5,000 finely annotated
frames, of which 2,975 are to be used for training, 500 for validation and the
rest for testing. Each finely annotated frame belongs to a 1.8-second-long
video clip with 30 frames, of which the 20\textsuperscript{th} frame is
annotated. The remaining 29 frames are not annotated. In addition, it also
provides 20,000 coarsely annotated frames. Each frame has a resolution of
\(1920\times1080\). \Cref{fig:cityscapes_annotation_example} shows examples of
fine-grained and coarse annotations provided by the dataset. Most semantic
segmentation models in the literature only use the 2,975 fine-grained
annotations for training purposes, ignoring the 20,000 coarse annotations and
all unannotated frames. We also use the IDD~\citep{DBLP:conf/wacv/VarmaSNCJ19}
dataset to show some generality in our experiments involving unannotated
frames. Since the full IDD dataset involves mixed-resolution videos and is
quite large, we limit ourselves to the subset of the dataset, which includes
only \(1920\times1080\) resolution frames. Even with such subsampling, we end
up with 4,032 training and 441 validation samples, all of which are finely
annotated, along with about 20 times as many unannotated frames.

As for our semantic segmentation model, we perform most of our experiments
using TMANet proposed by \citet{DBLP:conf/icip/WangW021} as we found it to be
one of the more recent video segmentation models which performs well on the
Cityscapes dataset. The authors introduce and use temporal memory attention to
attend to and integrate features from past frames while predicting the
segmentation mask for the current frame. While the authors use up to 4 past
frames, our experiments use a single past frame due to resource constraints. We
also use TDNet as proposed by \citet{DBLP:conf/cvpr/HuCWLSP20} for our
successful experiments to verify whether the approaches are applicable across
other video segmentation models. TDNet extracts image features using
subnetworks distributed over time and aggregates these features using the
attention propagation module to segment the current frame. We use
TD\textsuperscript{2}-FANet18, which uses two FANet18 subnetworks and works
with two past frames. Both networks are trained for 40,000 iterations for all
experiments.

\section[Method]{Method \footnote{\url{https://github.com/samiksome92/samsam2}}}
\label{sec:method}
We approach the problem of reducing annotation effort in two primary ways.
First, by trying to generate pseudo-labels for unannotated frames for provided
video clips, and second, by trying to use and refine coarse annotations
provided by the datasets. Our goal is to utilise foundation models, SAM and SAM
2, for these tasks.

\begin{figure*}[t]
    \centering
    \begin{subfigure}{0.33\linewidth}
        \includegraphics[width=\linewidth]{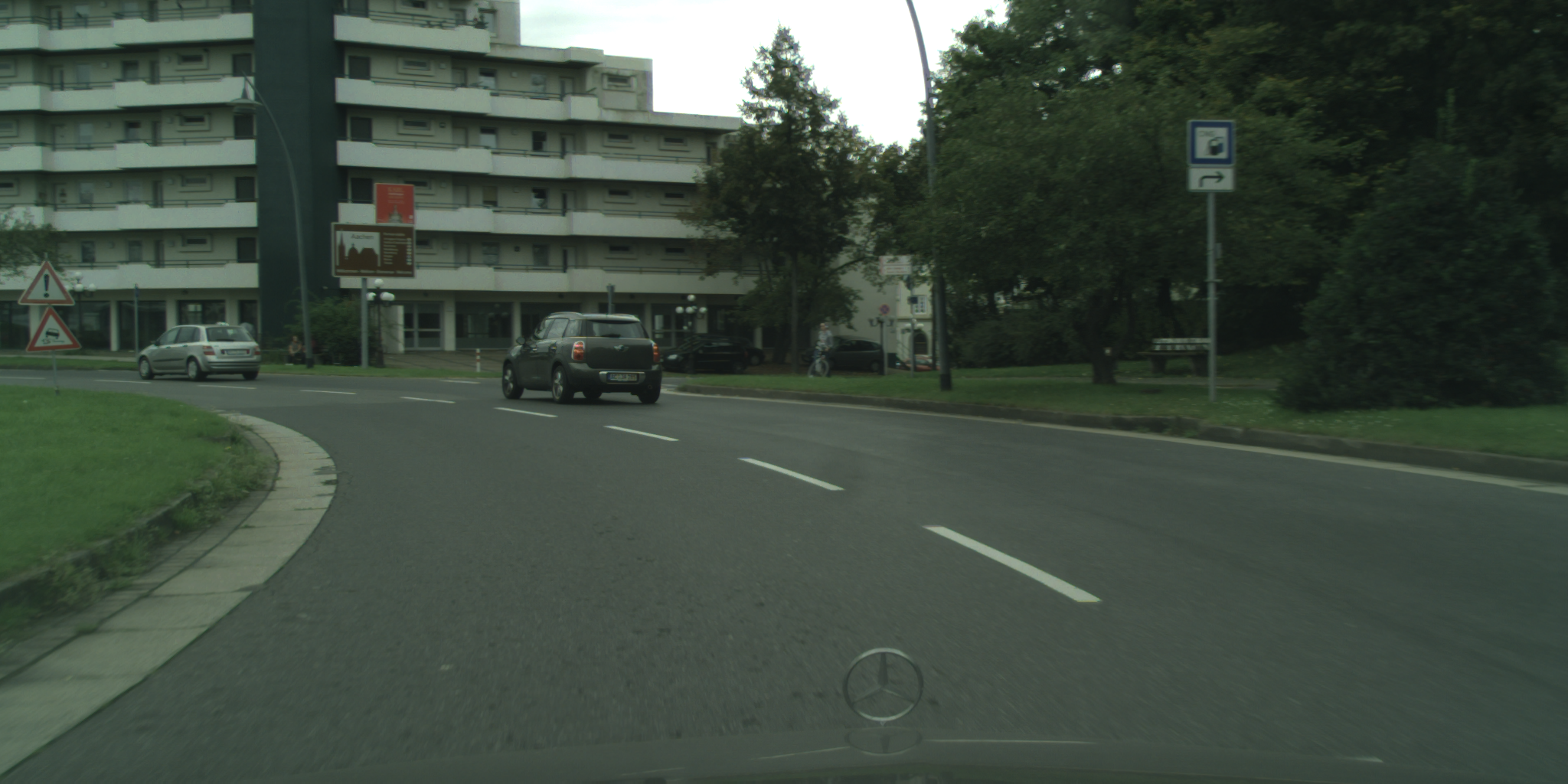}
        \caption{10\textsuperscript{th} Frame - Image}
    \end{subfigure}
    \begin{subfigure}{0.33\linewidth}
        \includegraphics[width=\linewidth]{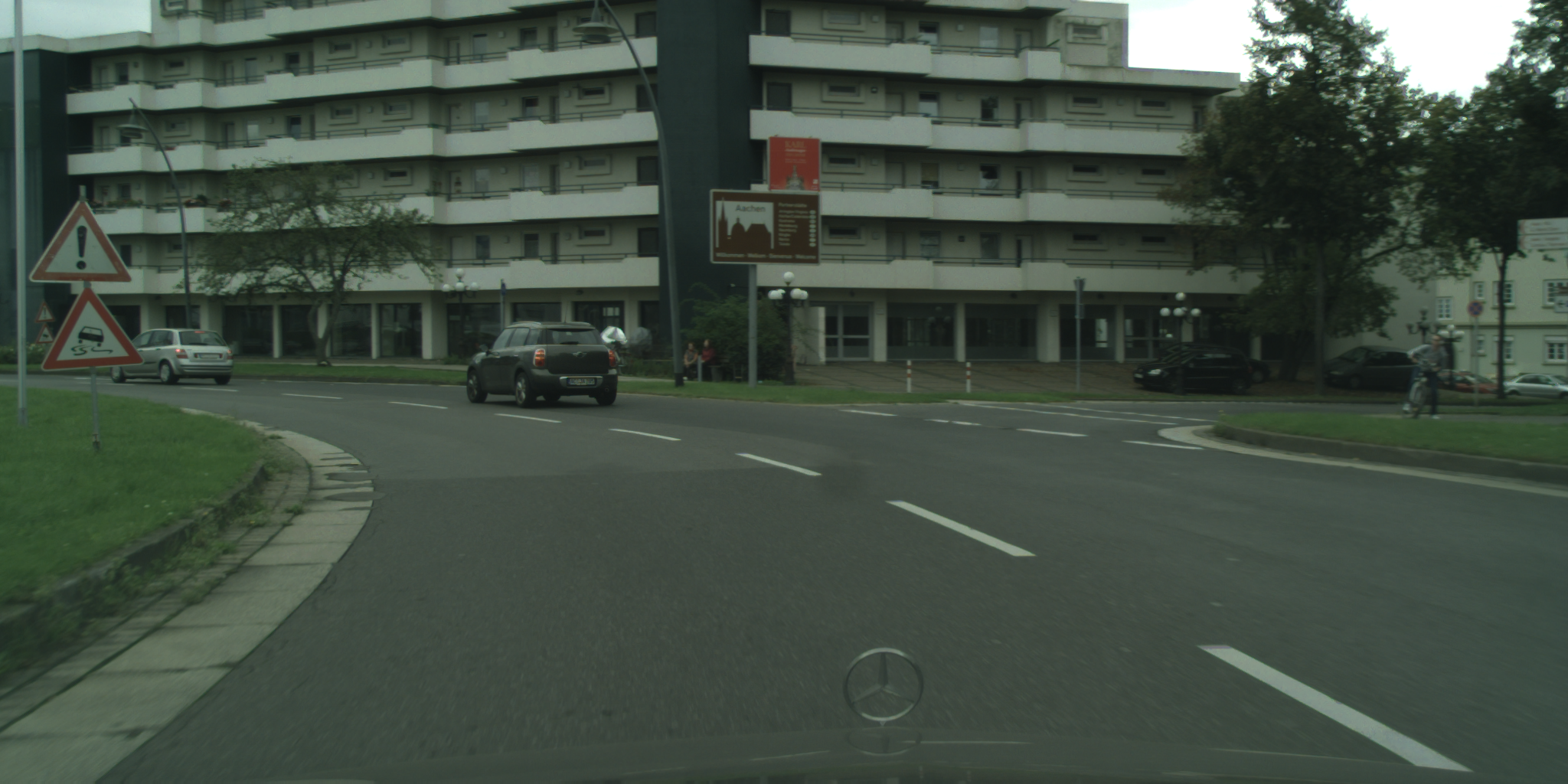}
        \caption{20\textsuperscript{th} Frame - Image}
    \end{subfigure}
    \begin{subfigure}{0.33\linewidth}
        \includegraphics[width=\linewidth]{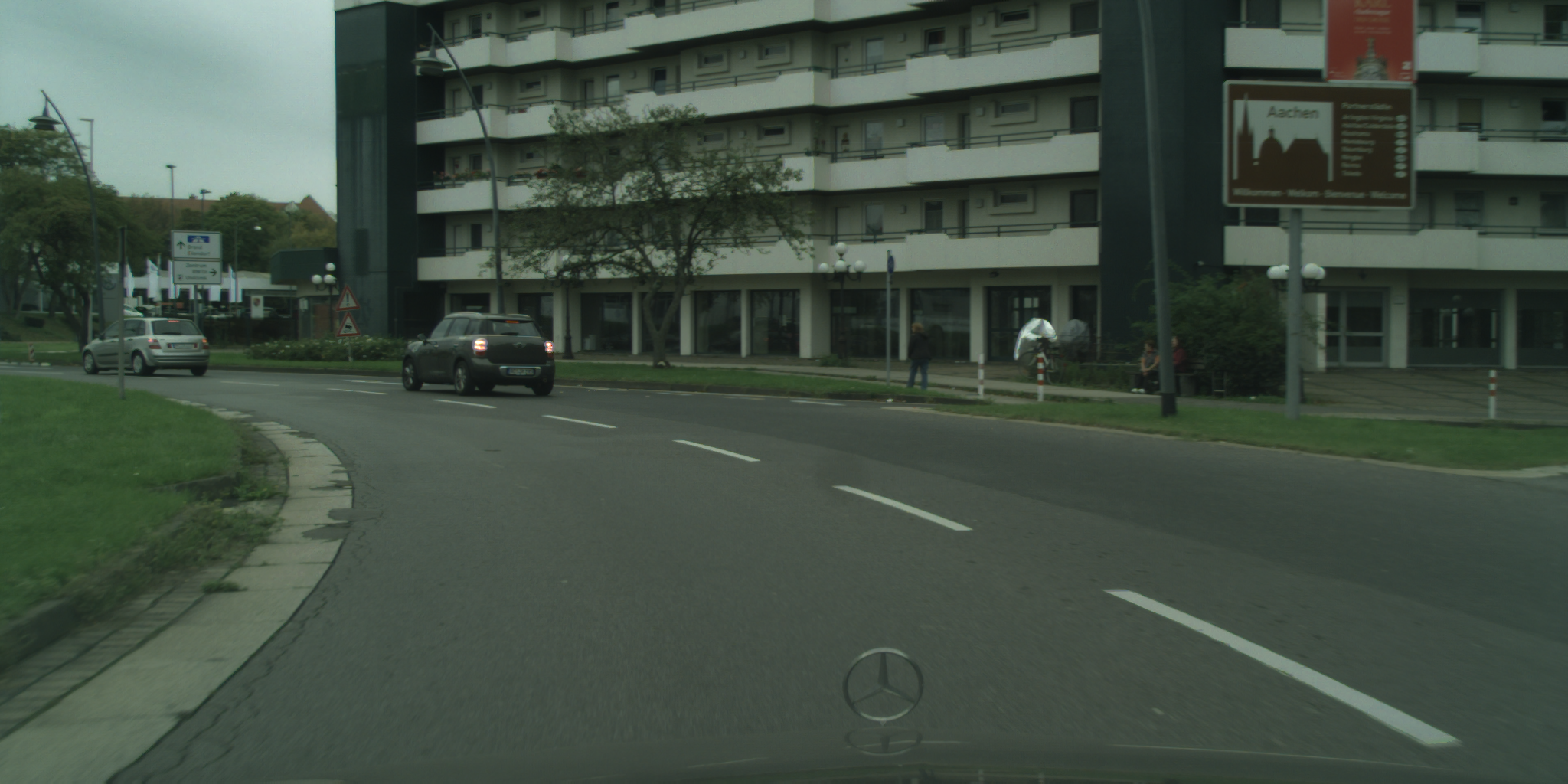}
        \caption{30\textsuperscript{th} Frame - Image}
    \end{subfigure}

    \begin{subfigure}{0.33\linewidth}
        \includegraphics[width=\linewidth]{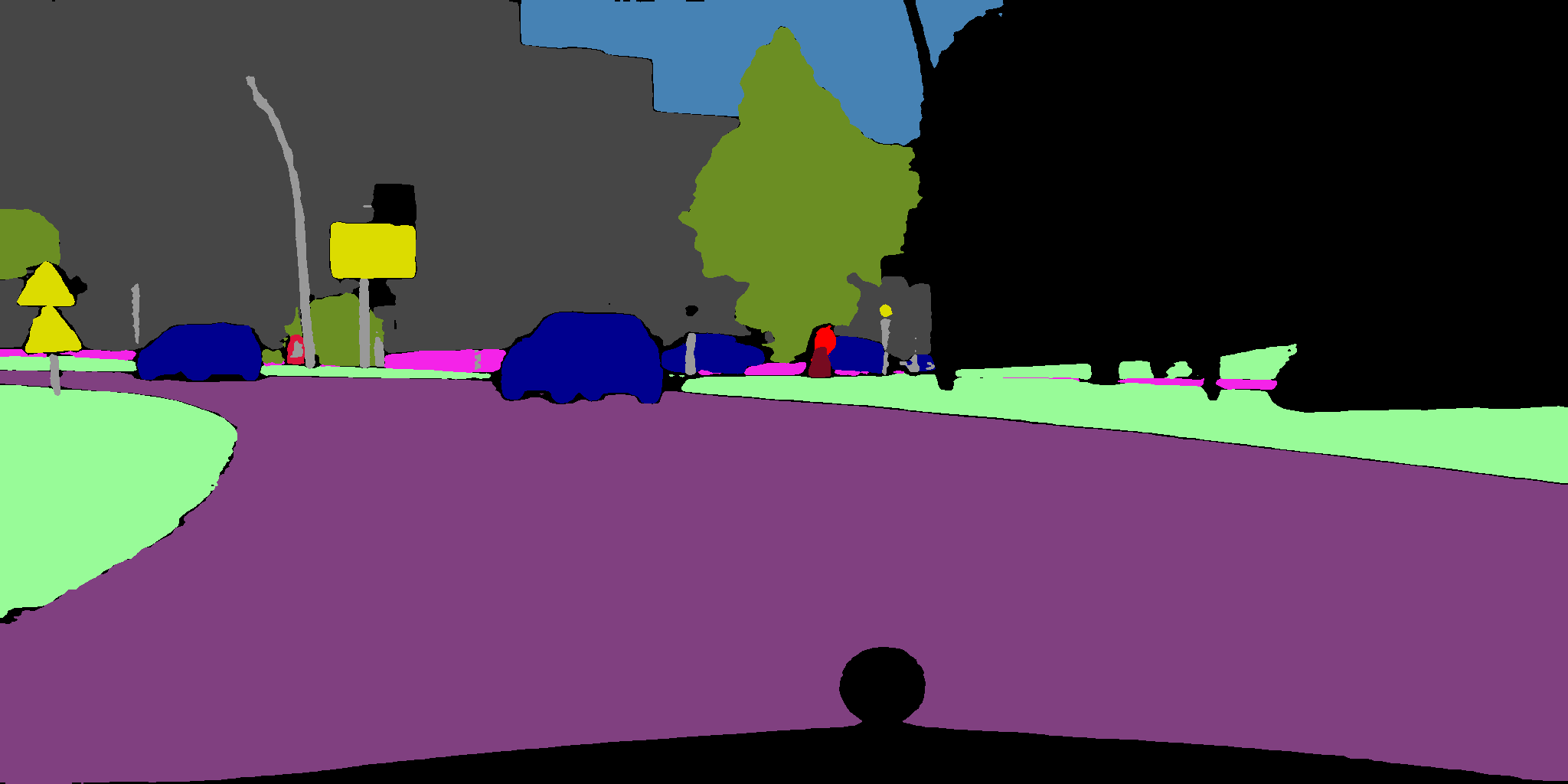}
        \caption{10\textsuperscript{th} Frame - Predicted Mask}
    \end{subfigure}
    \begin{subfigure}{0.33\linewidth}
        \includegraphics[width=\linewidth]{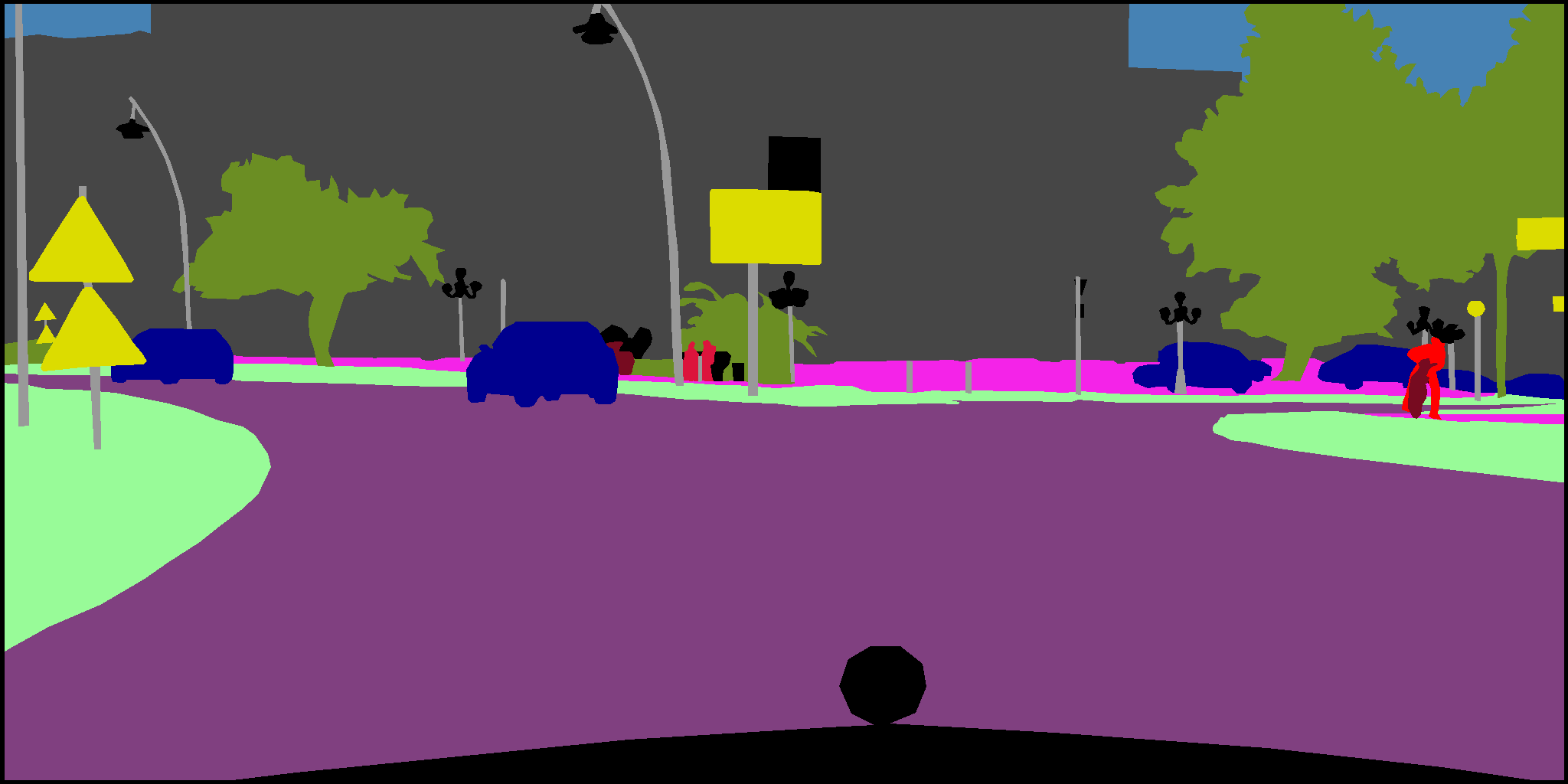}
        \caption{20\textsuperscript{th} Frame - Mask}
    \end{subfigure}
    \begin{subfigure}{0.33\linewidth}
        \includegraphics[width=\linewidth]{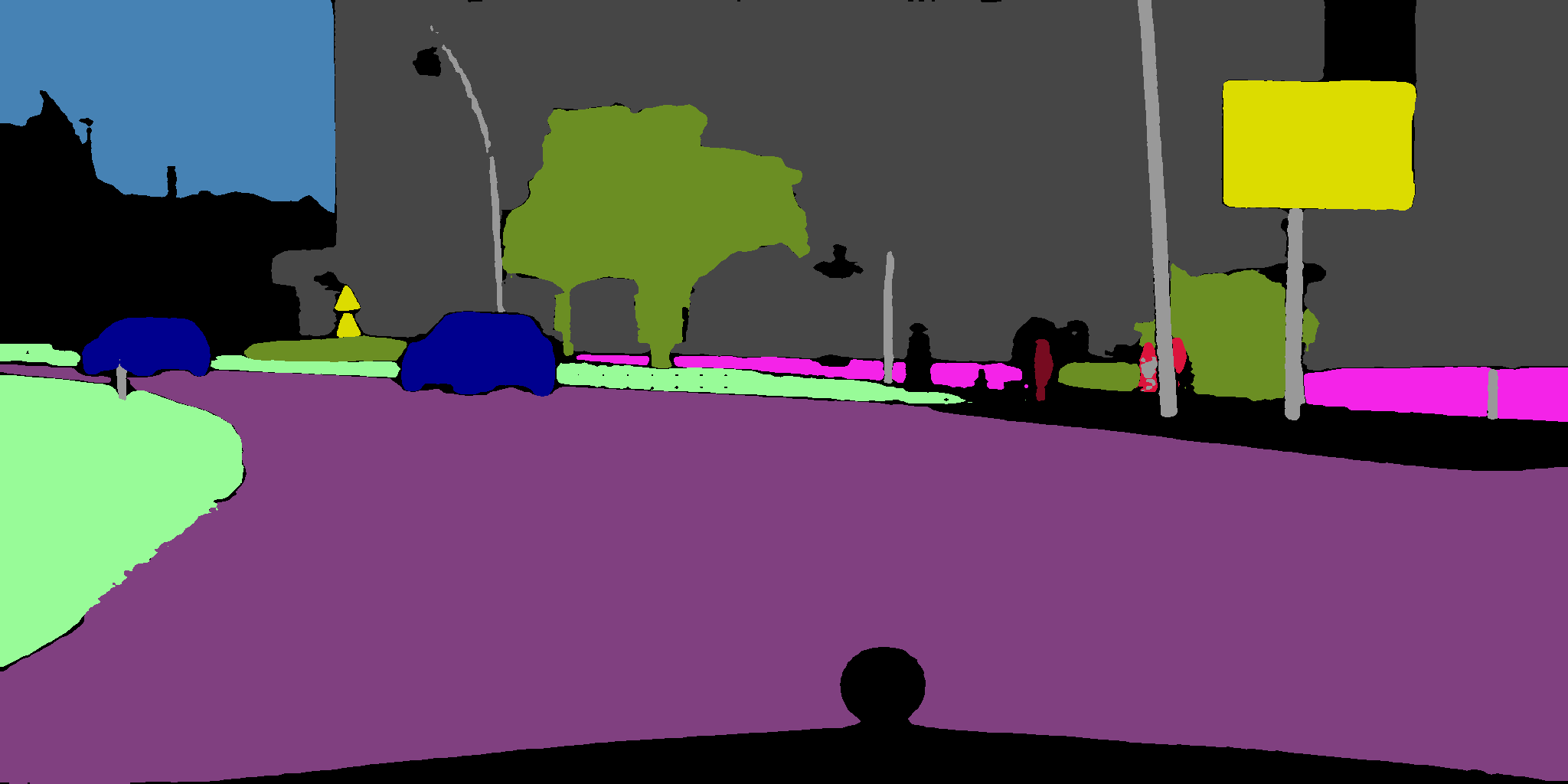}
        \caption{30\textsuperscript{th} Frame - Predicted Mask}
    \end{subfigure}
    \caption{An example showing how well SAM 2 can predict masks for past and future frames,
        given a manually annotated frame. Here, the 20\textsuperscript{th} frame is
        manually annotated. Masks for both 10\textsuperscript{th} and
        30\textsuperscript{th} frames were generated using SAM 2. We observe that SAM 2
        can track and propagate the masks for objects present in the
        20\textsuperscript{th} frame very well. However, it cannot, by design, tell us
        anything about new objects introduced in frames 10 and 30.} \Description{An
        example showing how well SAM 2 can predict masks for past and future frames,
        given a manually annotated frame. Here, the 20\textsuperscript{th} frame is
        manually annotated. Masks for both 10\textsuperscript{th} and
        30\textsuperscript{th} frames were generated using SAM 2. We observe that SAM 2
        can track and propagate the masks for objects present in the
        20\textsuperscript{th} frame very well. However, it cannot, by design, tell us
        anything about new objects introduced in frames 10 and 30.}
    \label{fig:no_annotations_example}
\end{figure*}

For the first task, generating labels for unannotated frames, we utilise
Segment Anything Model 2 (SAM 2). We take the following steps to generate
pseudo-labels via SAM 2.
\begin{enumerate}
    \item Set the available manual annotation for the 20\textsuperscript{th} frame as the
          initial segmentation mask for SAM 2.
    \item Run SAM 2 to track and generate segmentation masks for future frames.
    \item Reverse the frame order and run SAM 2 again to generate masks for past frames
\end{enumerate}
For simplicity, we only track up to 10 frames in each direction. Thus, after
SAM 2 is done predicting, we have pseudo-labels for frames 10 through 30. We
can now use these generated annotations to train our video segmentation models.
\Cref{fig:no_annotations_example} shows an example of how well SAM 2 is able to
track and predict masks for past and future frames.

\begin{figure*}[t]
    \centering
    \begin{subfigure}{0.33\linewidth}
        \includegraphics[width=\linewidth]{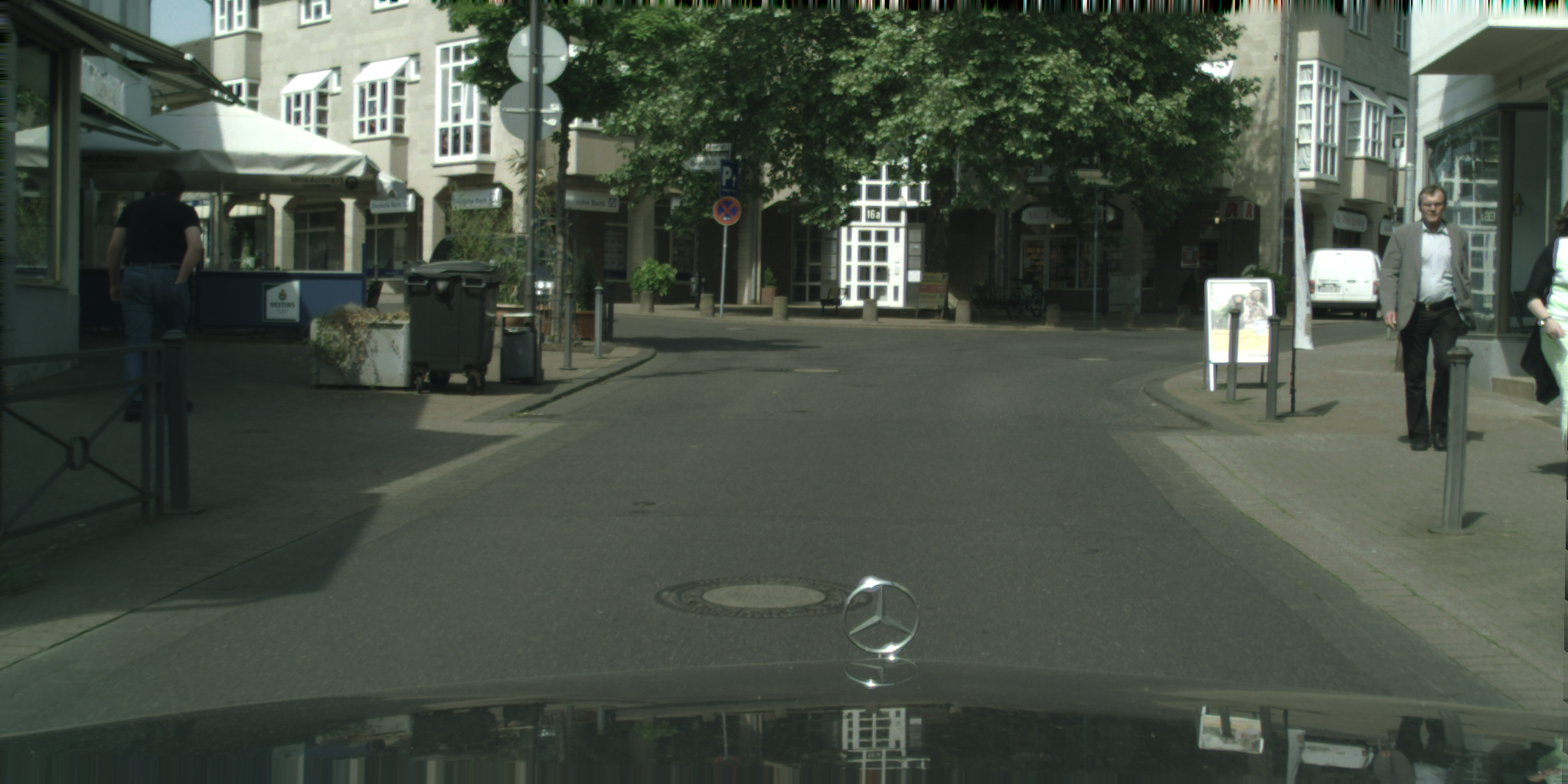}
        \caption{Image}
    \end{subfigure}
    \begin{subfigure}{0.33\linewidth}
        \includegraphics[width=\linewidth]{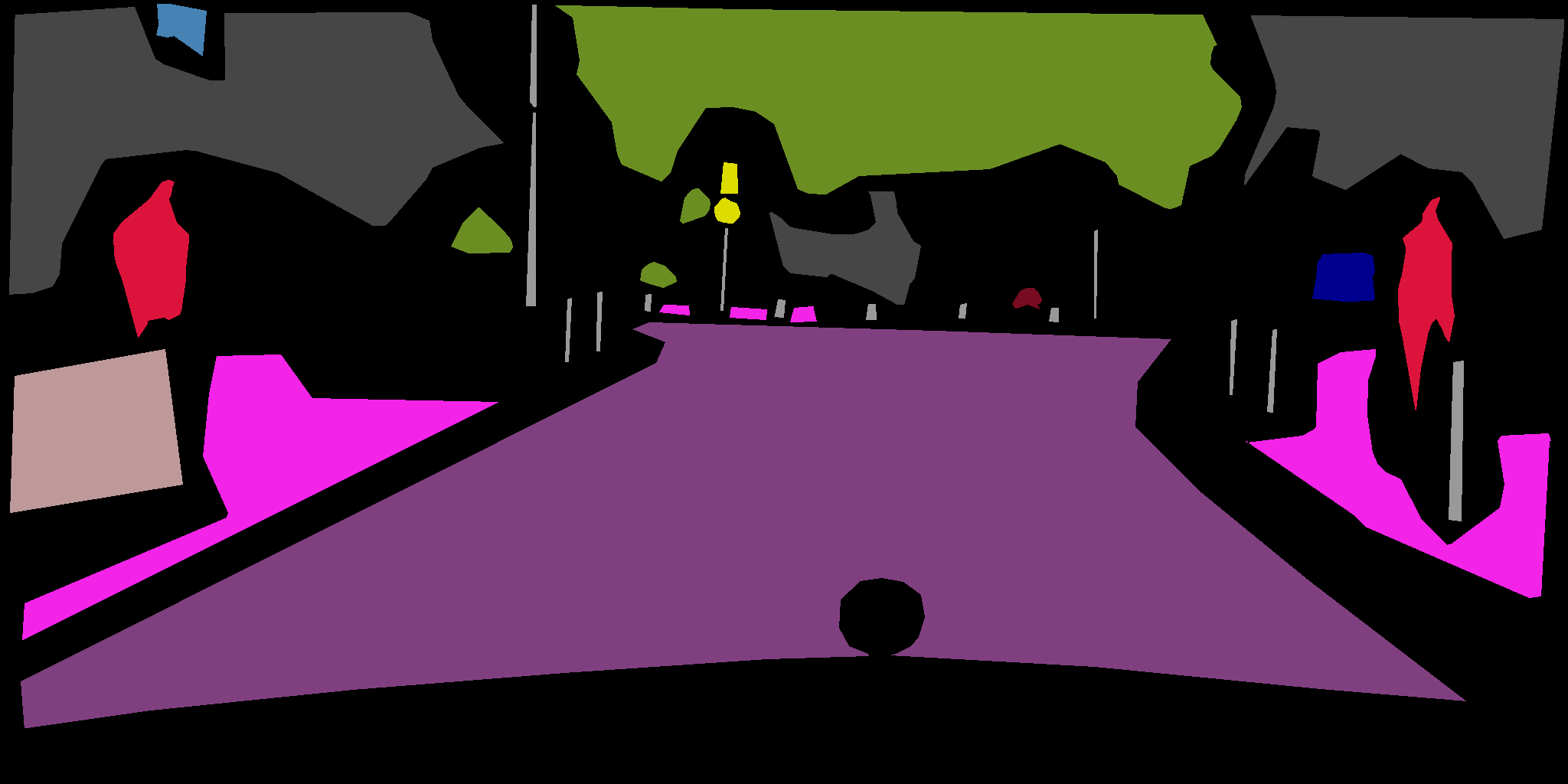}
        \caption{Coarse Mask}
    \end{subfigure}
    \begin{subfigure}{0.33\linewidth}
        \includegraphics[width=\linewidth]{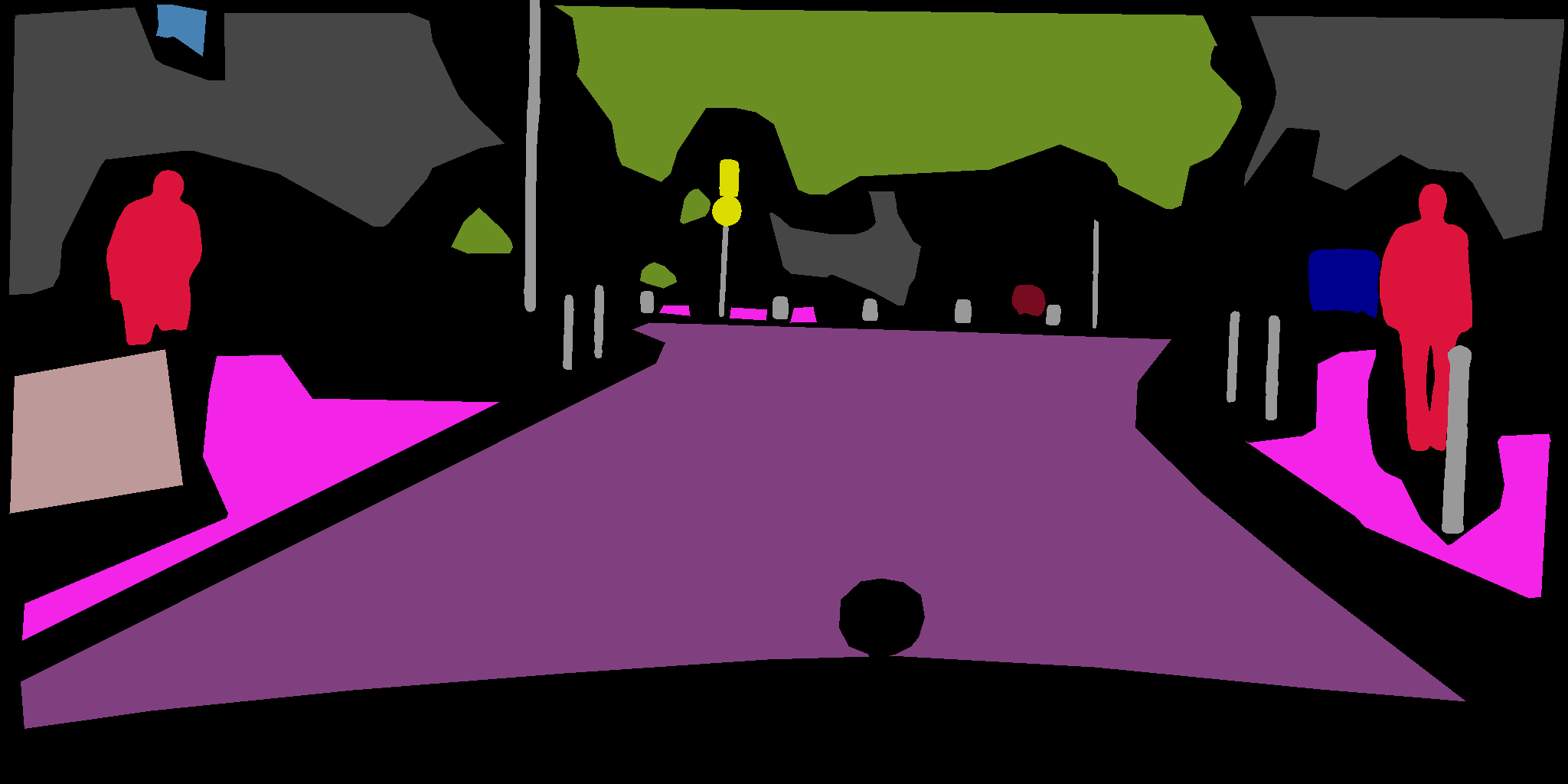}
        \caption{Refined Coarse Mask}
    \end{subfigure}

    \begin{subfigure}{0.33\linewidth}
        \includegraphics[width=\linewidth]{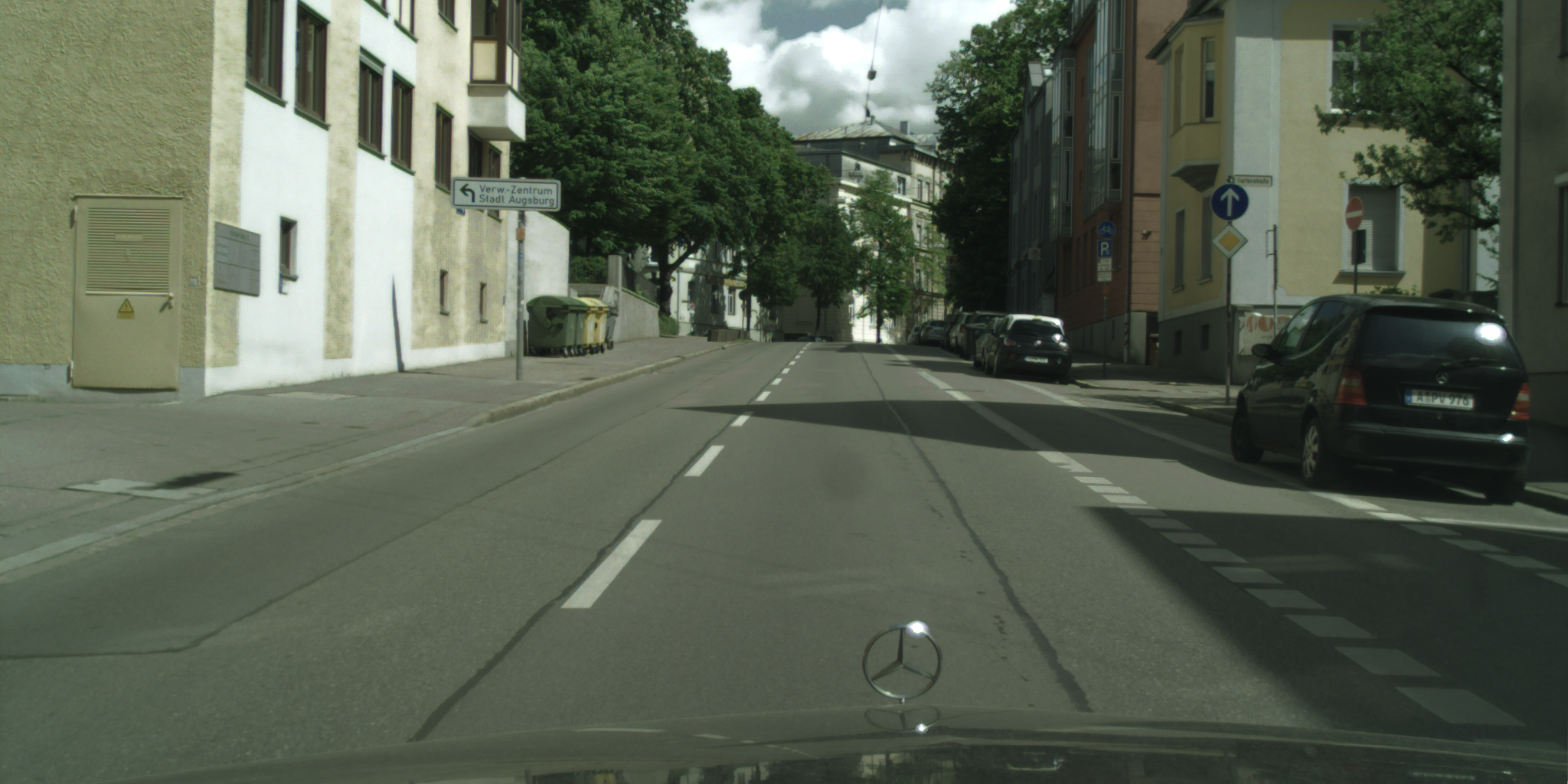}
        \caption{Image}
    \end{subfigure}
    \begin{subfigure}{0.33\linewidth}
        \includegraphics[width=\linewidth]{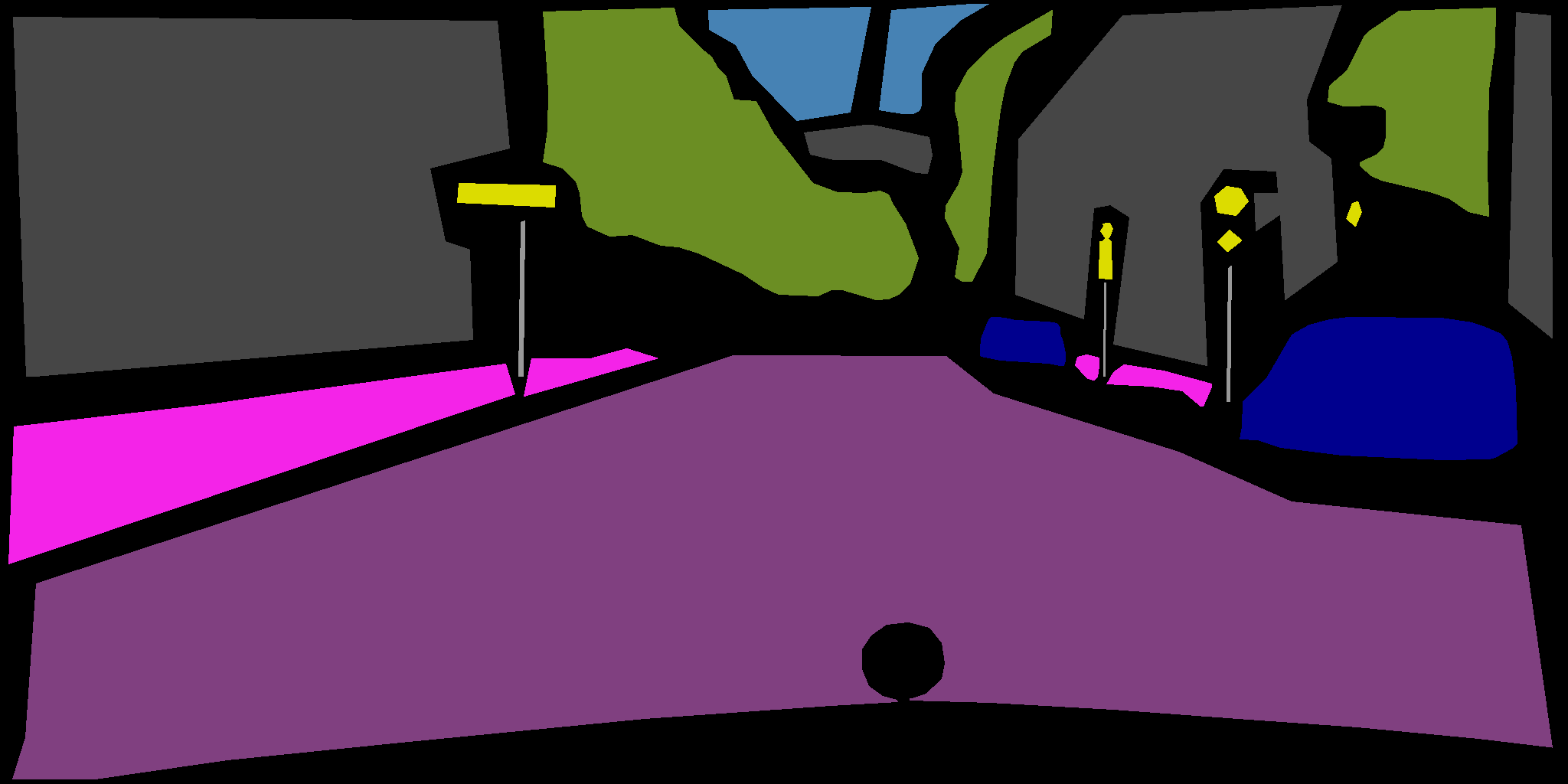}
        \caption{Coarse Mask}
    \end{subfigure}
    \begin{subfigure}{0.33\linewidth}
        \includegraphics[width=\linewidth]{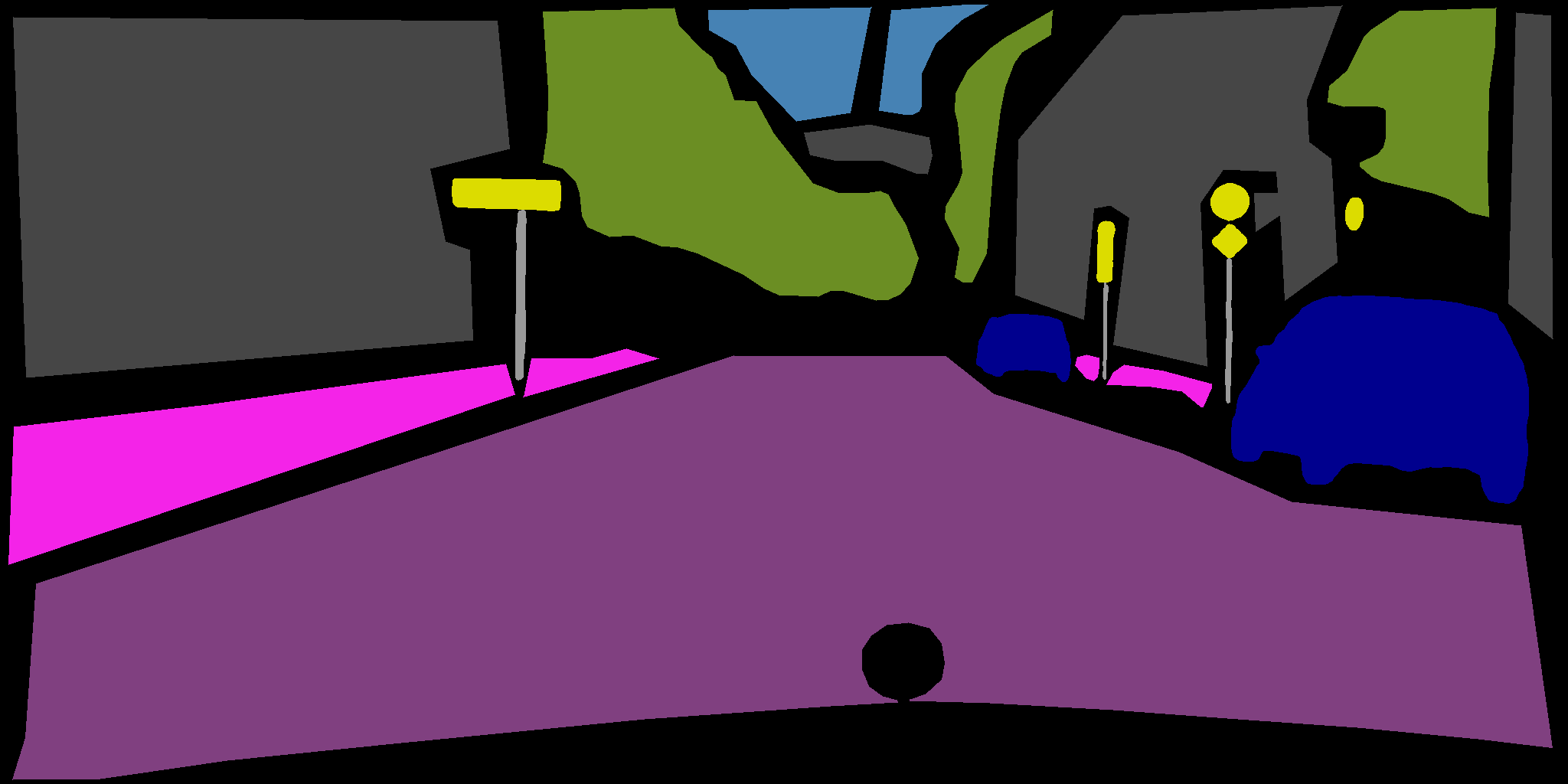}
        \caption{Refined Coarse Mask}
    \end{subfigure}
    \caption{A couple of examples from the Cityscapes dataset demonstrating how well SAM
        refines some of the segmentation classes. As can be seen, masks for objects
        like cars, traffic signs, poles and people have been refined quite nicely and
        can be treated as fine-grained instead of being only a rough approximation.}
    \Description{A couple of examples from the Cityscapes dataset demonstrating how
        well SAM refines some of the segmentation classes. As can be seen, masks for
        objects like cars, traffic signs, poles and people have been refined quite
        nicely and can be treated as fine-grained instead of being only a rough
        approximation.} \label{fig:coarse_annotations_sam_example}
\end{figure*}

For the second task, we need to refine coarse annotations. The Segment Anything
Model (SAM) can work with a variety of input options. After a few experiments,
we found that the following procedure worked well. For each instance level
segmentation mask:
\begin{enumerate}
    \item Randomly choose two points within the mask.
    \item Ask SAM to generate segmentation mask from the chosen points.
    \item Refine output mask using SAM for two iterations.
\end{enumerate}
We only refine masks for 8 out of 19 segmentation classes, namely, pole,
traffic light, traffic sign, person, rider, car, motorcycle and bicycle, and
keep the rest as is. For the other classes, we already have a sufficient number
of samples that are annotated as the classes, such as road and sky, are highly
prevalent in all frames. We discuss this further in
\cref{sec:coarse_annotations_what_does_not}. We show a couple of examples of
such refinement for the Cityscapes dataset in
\cref{fig:coarse_annotations_sam_example}.

\section{Experiments}
\label{sec:experiments}
Our experiments can be divided into two broad categories: Experiments with
unannotated frames and experiments involving coarsely annotated frames. In both
cases, our goal is to generate fine-grained pseudo-labels, which can be used to
augment existing data for training segmentation models. We aim to show that
instead of using all fine-grained manually annotated data, SAM and SAM 2 can be
used to generate more data, reducing the amount of manually annotated data
required with minimal loss of performance.

\begin{table*}[t]
    \centering
    \caption{Mean Intersection over Union (mIoU) for training experiments on the Cityscapes
        dataset using manually annotated and SAM 2-generated labels mixed together. The
        Manual \% column indicates what percentage of the training data was manually
        annotated. Filled boxes indicate frames that were used, whereas unfilled ones
        indicate unused frames. Green indicates manually annotated and black indicates
        SAM 2-generated masks.}
    \begin{tabular}{lccrr}
        \toprule
        ID  & Frames Used                                            & mIoU  & Manual \% & Total samples \\
            & 10 \S{0.085} 15 \S{0.085} 20 \S{0.085} 25 \S{0.085} 30 &       &           &               \\
        \midrule
        \multicolumn{5}{c}{TMANet}                                                                       \\
        N1  & \W\W\W\W\W\W\W\W\W\W\k\W\W\W\W\W\W\W\W\W\W             & 75.65 & 100.00    & 2975          \\
        P2  & \K\W\W\W\W\W\W\W\W\W\k\W\W\W\W\W\W\W\W\W\W             & 75.75 & 50.00     & 2976          \\
        P2s & \W\W\W\W\W\W\W\W\W\K\k\W\W\W\W\W\W\W\W\W\W             & 74.83 & 50.00     & 2976          \\
        F2  & \W\W\W\W\W\W\W\W\W\W\k\W\W\W\W\W\W\W\W\W\K             & 75.11 & 50.00     & 2976          \\
        F2s & \W\W\W\W\W\W\W\W\W\W\k\K\W\W\W\W\W\W\W\W\W             & 74.16 & 50.00     & 2976          \\
        B2n & \K\W\W\W\W\W\W\W\W\W\w\W\W\W\W\W\W\W\W\W\K             & 74.84 & 0.00      & 2976          \\
        B3  & \K\W\W\W\W\W\W\W\W\W\k\W\W\W\W\W\W\W\W\W\K             & 74.48 & 33.33     & 2976          \\
        B3s & \W\W\W\W\W\W\W\W\W\K\k\K\W\W\W\W\W\W\W\W\W             & 68.79 & 33.33     & 2976          \\
        B5  & \K\W\W\W\W\K\W\W\W\W\k\W\W\W\W\K\W\W\W\W\K             & 72.90 & 20.00     & 2975          \\
        B5s & \W\W\W\W\W\W\W\W\K\K\k\K\K\W\W\W\W\W\W\W\W             & 72.31 & 20.00     & 2975          \\
        B7  & \K\W\W\K\W\W\W\K\W\W\k\W\W\K\W\W\W\K\W\W\K             & 69.68 & 14.29     & 2975          \\
        B9  & \K\W\W\K\W\K\W\K\W\W\k\W\W\K\W\K\W\K\W\W\K             & 64.07 & 11.11     & 2979          \\
        B11 & \K\W\K\W\K\W\K\W\K\W\k\W\K\W\K\W\K\W\K\W\K             & 63.70 & 9.09      & 2981          \\
        P11 & \K\K\K\K\K\K\K\K\K\K\k\W\W\W\W\W\W\W\W\W\W             & 65.98 & 9.09      & 2981          \\
        F11 & \W\W\W\W\W\W\W\W\W\W\k\K\K\K\K\K\K\K\K\K\K             & 64.07 & 9.09      & 2981          \\
        \midrule
        \multicolumn{5}{c}{TDNet}                                                                        \\
        N1  & \W\W\W\W\W\W\W\W\W\W\k\W\W\W\W\W\W\W\W\W\W             & 71.52 & 100.00    & 2975          \\
        B3  & \K\W\W\W\W\W\W\W\W\W\k\W\W\W\W\W\W\W\W\W\K             & 69.41 & 33.33     & 2976          \\
        B5  & \K\W\W\W\W\K\W\W\W\W\k\W\W\W\W\K\W\W\W\W\K             & 66.36 & 20.00     & 2975          \\
        \bottomrule
    \end{tabular}
    \label{tab:no_annotations}
\end{table*}

\begin{table*}[t]
    \centering
    \caption{Mean Intersection over Union (mIoU) for training experiments on the IDD dataset
        using manually annotated and SAM 2-generated labels mixed together. The Manual
        \% column indicates what percentage of the training data was manually
        annotated. Filled boxes indicate frames that were used, whereas unfilled ones
        indicate unused frames. Green indicates manually annotated and black indicates
        SAM 2-generated masks.}
    \begin{tabular}{lccrr}
        \toprule
        ID & Frames Used                                                & mIoU  & Manual \% & Total samples \\
           & -10 \S{0.09} -5 \S{0.09} 0 \S{0.09} 5 \S{0.09} 10 \S{0.01} &       &           &               \\
        \midrule
        N1 & \W\W\W\W\W\W\W\W\W\W\k\W\W\W\W\W\W\W\W\W\W                 & 62.15 & 100.00    & 4032          \\
        F2 & \W\W\W\W\W\W\W\W\W\W\k\W\W\W\W\W\W\W\W\W\K                 & 60.67 & 50.00     & 4032          \\
        B3 & \K\W\W\W\W\W\W\W\W\W\k\W\W\W\W\W\W\W\W\W\K                 & 61.31 & 33.33     & 4032          \\
        \bottomrule
    \end{tabular}
    \label{tab:no_annotations_idd}
\end{table*}

\subsection{Unannotated Frames}
\label{sec:no_annotations}
First, we look at ways in which we can utilise SAM 2 to generate annotations
for frames that are not annotated at all. We use the training set with 2,975
clips for this purpose. Each of these clips has 30 frames, of which the
20\textsuperscript{th} frame is finely annotated. We utilise SAM 2 as stated in
\cref{sec:method} to generate labels for the remaining frames. In the following
sections, we discuss experiments that produced good results as well as others
that did not work out well.

\subsubsection{Successful Approaches}
First, we discuss a couple of approaches that work well. We replace half of the
finely annotated frames in the dataset with a predicted frame from the same
clip. We consider both past and future frames for our experiments. However, we
only use the 10\textsuperscript{th} and 30\textsuperscript{th} frames.
Essentially, the frames that are furthest in time from the manually annotated
20\textsuperscript{th} frame. \Cref{tab:no_annotations} shows the results of
these experiments on the Cityscapes dataset as \textbf{P2} and \textbf{F2},
respectively. For comparison, \textbf{N1} is when we train with all manually
annotated frames.

From the table, it is clear that using only half of the manually annotated
frames, along with predictions from SAM 2, provides nearly the same results as
using all annotated frames. However, this reduces the requirement for manual
annotation by half. We take this even further by keeping only one-third of the
manually annotated frames and using both the 10\textsuperscript{th} and
30\textsuperscript{th} frames to replace the rest. The result is shown in
\cref{tab:no_annotations} as \textbf{B3}. Once again, the mean Intersection
over Union (mIoU) is only a little less than \textbf{N1}, whereas the
annotation effort required has now become one-third of the original.

We also show the results of experiments with the TDNet model in the same table.
Since TDNet requires much longer training time compared to TMANet, we were only
able to test a few scenarios, \textbf{N1}, \textbf{B3} and \textbf{B5}.

Similarly, we also show results with the TMANet model on the IDD dataset to
show some generality across datasets in \cref{tab:no_annotations_idd}. Once
again, we see that in both \textbf{F2} and \textbf{B3}, the drop in performance
is minimal compared to the reduction of effort required for manual annotation.

\begin{figure}
    \centering
    \includegraphics[width=\linewidth]{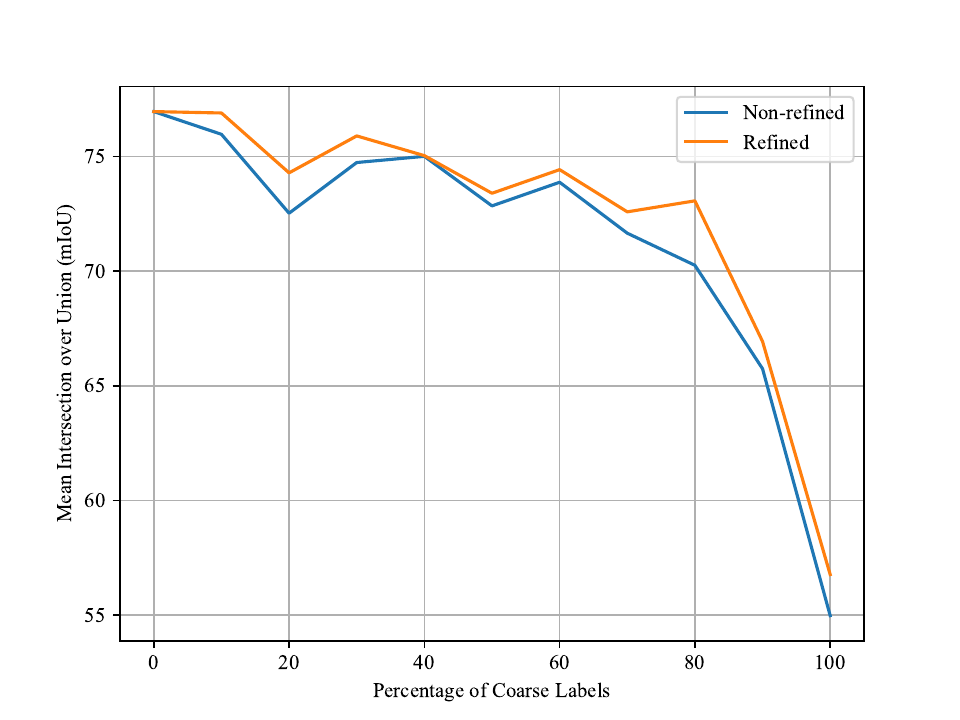}
    \caption{Plot showing changes in mean Intersection over Union (mIoU) against increasing
        percentage of coarse annotations in the training data for TMANet on the
        Cityscapes dataset. The non-refined and refined values refer to the coarse
        annotations being used as-is versus being refined using SAM.} \Description{Plot
        showing changes in mean Intersection over Union (mIoU) against increasing
        percentage of coarse annotations in the training data for TMANet on the
        Cityscapes dataset. The non-refined and refined values refer to the coarse
        annotations being used as-is versus being refined using SAM.}
    \label{fig:coarse_annotations_sam_miou}
\end{figure}

\begin{table}[t]
    \centering
    \caption{Mean IoU and Accuracy for increasing ratio of non-refined and refined coarse
        annotations with TMANet on the Cityscapes dataset. The Coarse \% column
        indicates the percentage of coarse annotations in the training samples.}
    \label{tab:coarse_annotations_tmanet}
    \begin{tabular}{Sccccc}
        \toprule
        {Coarse \%} & \multicolumn{2}{c}{Non-refined} & \multicolumn{2}{c}{Refined} \\
                    & mIoU  & mAcc                    & mIoU  & mAcc                \\
        \midrule
        0           & 76.95 & 86.19                   & 76.95 & 86.19               \\
        10          & 75.96 & 84.68                   & 76.89 & 86.12               \\
        20          & 72.52 & 83.00                   & 74.28 & 84.03               \\
        30          & 74.73 & 83.63                   & 75.89 & 84.88               \\
        40          & 75.00 & 83.88                   & 75.03 & 84.39               \\
        50          & 72.84 & 83.35                   & 73.39 & 83.03               \\
        60          & 73.87 & 83.06                   & 74.42 & 84.52               \\
        70          & 71.65 & 80.04                   & 72.58 & 82.26               \\
        80          & 70.25 & 80.79                   & 73.06 & 83.13               \\
        90          & 65.74 & 75.23                   & 66.93 & 77.04               \\
        100         & 54.98 & 68.38                   & 56.77 & 69.73               \\
        \bottomrule
    \end{tabular}
\end{table}

\subsubsection{Failed Approaches}
We now turn our attention to several other approaches that were tried but did
not work well. The very first question that comes to mind after seeing the
results of \textbf{P2}, \textbf{F2}, \textbf{B3} in \cref{tab:no_annotations}
is, why stop at only two generated frames? Why not use all 20 generated frames?
That would reduce the annotation effort by 20 times. Unfortunately, it does not
work well, as evident by \textbf{B7}, \textbf{B9}, \textbf{B11}, etc. in the
table. We believe the reason for this is that the frames included are very
similar to each other, and not simply because of the quality of the predicted
labels. This is further supported by the results of \textbf{B2n}, where we only
train on generated labels (for 10\textsuperscript{th} and
30\textsuperscript{th} frames) and yet see similar performance to \textbf{N1}.
If the quality of the generated labels had not been good, this scenario would
have performed much more poorly, as it contains no manually annotated data at
all. To test this hypothesis further, we also performed experiments where all
frames are sequential, such as \textbf{P2s}, \textbf{F2s}, \textbf{B3s},
\textbf{B5s}. We observe that the results are always worse than their more
distributed counterparts \textbf{P2}, \textbf{F2}, \textbf{B3}, \textbf{B5},
further indicating that diversity among training samples is important.

In order to leverage the importance of diversity, we tried several experiments
where, instead of randomly selecting subsets of clips for training, we tried to
select clips which had the highest diversity between frames. We used optical
flow between frames to see which clips generally had the highest flow
magnitudes, but were unsuccessful in producing better results than random
sampling. We also considered improving the masks produced by SAM 2. Since SAM 2
can only track and predict masks for objects present in the
20\textsuperscript{th} frame, any new object appearing in later frames is
necessarily unlabelled. We tried to fill these gaps by feeding these frames to
the \textbf{B3} model. These rectified frames would then be again used to train
a new model, essentially creating an iterative dog-fooding-based training
procedure. However, the results were not as good as the simple non-iterative
training.

\begin{figure}
    \centering
    \includegraphics[width=\linewidth]{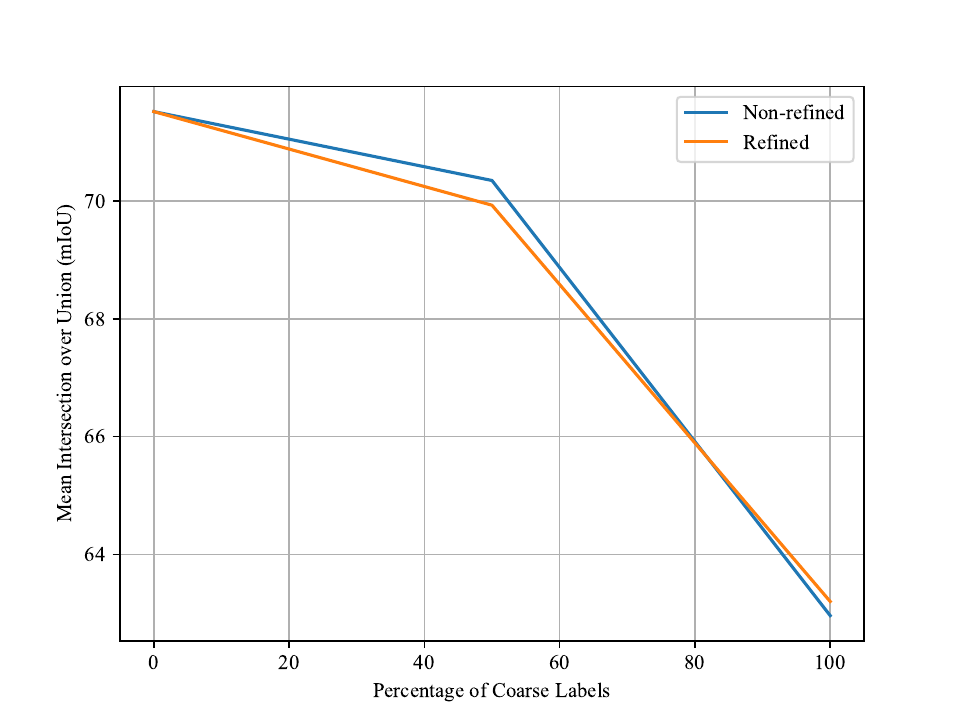}
    \caption{Plot showing changes in mean Intersection over Union (mIoU) against increasing
        percentage of coarse annotations in the training data for TDNet on the
        Cityscapes dataset. The non-refined and refined values refer to the coarse
        annotations being used as-is versus being refined using SAM.} \Description{Plot
        showing changes in mean Intersection over Union (mIoU) against increasing
        percentage of coarse annotations in the training data for TDNet on the
        Cityscapes dataset. The non-refined and refined values refer to the coarse
        annotations being used as-is versus being refined using SAM.}
    \label{fig:coarse_annotations_sam_miou_tdnet}
\end{figure}

\begin{table}[t]
    \centering
    \caption{Mean IoU and Accuracy for increasing ratio of non-refined and refined coarse
        annotations with TDNet on the Cityscapes dataset. The Coarse \% column
        indicates the percentage of coarse annotations in the training samples.}
    \label{tab:coarse_annotations_tdnet}
    \begin{tabular}{Sccccc}
        \toprule
        {Coarse \%} & \multicolumn{2}{c}{Non-refined} & \multicolumn{2}{c}{Refined} \\
                    & mIoU  & mAcc                    & mIoU  & mAcc                \\
        \midrule
        0           & 71.52 & 79.46                   & 71.52 & 79.46               \\
        50          & 70.35 & 78.56                   & 69.93 & 78.26               \\
        100         & 62.96 & 73.86                   & 63.20 & 75.42               \\
        \bottomrule
    \end{tabular}
\end{table}

\begin{table*}[t]
    \centering
    \caption{Classwise Mean Intersection over Union for a few mixtures with TMANet on the
        Cityscapes dataset, for both non-refined and refined annotations.}
    \begin{tabular}{lccccccccccccc}
        \toprule
        Class             & 0\% Coarse &  & \multicolumn{2}{c}{20\% Coarse} &  & \multicolumn{2}{c}{50\% Coarse} &  & \multicolumn{2}{c}{80\% Coarse} &  & \multicolumn{2}{c}{100\% Coarse} \\
                          &            &  & NR    & R                       &  & NR    & R                       &  & NR    & R                       &  & NR    & R                        \\
        \midrule
        road              & 97.90      &  & 97.81 & 97.85                   &  & 97.85 & 97.89                   &  & 97.52 & 97.68                   &  & 95.48 & 95.78                    \\
        sidewalk          & 84.07      &  & 82.89 & 83.42                   &  & 83.17 & 83.46                   &  & 80.79 & 82.09                   &  & 69.04 & 71.03                    \\
        building          & 92.52      &  & 91.48 & 91.87                   &  & 91.30 & 91.52                   &  & 90.39 & 90.90                   &  & 85.81 & 86.29                    \\
        wall              & 62.24      &  & 55.68 & 54.54                   &  & 52.53 & 49.19                   &  & 50.47 & 53.74                   &  & 37.19 & 37.51                    \\
        fence             & 61.79      &  & 57.07 & 57.59                   &  & 56.52 & 56.59                   &  & 53.08 & 54.81                   &  & 33.69 & 39.01                    \\
        \B{pole}          & 63.25      &  & 59.70 & 61.29                   &  & 59.85 & 61.34                   &  & 57.17 & 58.84                   &  & 44.55 & 44.85                    \\
        \B{traffic light} & 70.02      &  & 65.35 & 67.82                   &  & 65.68 & 67.29                   &  & 63.10 & 65.15                   &  & 51.47 & 55.51                    \\
        \B{traffic sign}  & 77.91      &  & 73.96 & 75.43                   &  & 73.31 & 76.40                   &  & 71.73 & 73.79                   &  & 61.56 & 65.17                    \\
        vegetation        & 92.49      &  & 91.87 & 91.96                   &  & 91.66 & 91.73                   &  & 90.91 & 91.25                   &  & 86.41 & 86.72                    \\
        terrain           & 65.28      &  & 62.20 & 63.29                   &  & 60.43 & 59.57                   &  & 58.12 & 60.36                   &  & 46.37 & 45.10                    \\
        sky               & 94.37      &  & 93.88 & 93.68                   &  & 93.96 & 93.73                   &  & 92.98 & 93.61                   &  & 90.37 & 91.09                    \\
        \B{person}        & 81.67      &  & 78.43 & 79.66                   &  & 77.30 & 76.94                   &  & 75.13 & 76.93                   &  & 57.76 & 60.91                    \\
        \B{rider}         & 62.66      &  & 53.16 & 58.54                   &  & 56.14 & 53.71                   &  & 45.53 & 52.59                   &  & 26.39 & 30.94                    \\
        \B{car}           & 94.98      &  & 94.01 & 94.26                   &  & 93.85 & 94.03                   &  & 93.09 & 93.21                   &  & 85.79 & 84.36                    \\
        truck             & 70.59      &  & 66.93 & 73.20                   &  & 66.52 & 72.67                   &  & 64.81 & 69.25                   &  & 39.51 & 55.17                    \\
        bus               & 77.67      &  & 72.24 & 74.98                   &  & 73.33 & 78.87                   &  & 73.89 & 80.54                   &  & 49.74 & 41.73                    \\
        train             & 68.60      &  & 55.70 & 62.59                   &  & 61.80 & 57.40                   &  & 61.58 & 72.18                   &  & 00.00 & 01.83                    \\
        \B{motorcycle}    & 67.39      &  & 52.16 & 54.94                   &  & 55.34 & 57.30                   &  & 43.84 & 48.86                   &  & 22.57 & 25.62                    \\
        \B{bicycle}       & 76.64      &  & 73.34 & 74.40                   &  & 73.42 & 74.80                   &  & 70.57 & 72.34                   &  & 60.86 & 59.97                    \\
        \bottomrule
    \end{tabular}
    \label{tab:classwise_miou}
\end{table*}

\subsection{Coarse Annotations}
So far, we have seen how SAM 2 can be used to generate pseudo-labels for
completely unlabelled past and future frames, given a manually annotated frame.
However, as stated earlier, the Cityscapes dataset also provides us with a
large number of coarse annotations. We also wanted to find ways to incorporate
these coarse samples during the training process to either improve or get
comparable performance to using only finely annotated samples. Moreover, if
possible, we also wanted to refine these annotations using SAM or SAM 2.

First, we discuss some experiments where we simply used the coarse frames as
is, without any refinement. We tried a couple of approaches. Initially, we
tried to train the model with only coarse samples, with the idea being that
since there are much larger number of coarse samples compared to fine-grained
samples (20,000 to 2,975), the model may learn to get comparable performance.
However, we did not achieve good results. We then tried mixing all 20,000
coarse annotations with the 2,975 fine annotations to see if we could achieve
better results. While we did get better results than using only coarse
annotations, the performance was much worse than when using only fine-grained
annotations.

We finally started obtaining reasonable results when we mixed coarse and fine
annotations in more equal ratios. With this knowledge, we perform a set of
experiments where we keep the number of training samples fixed at 2,975 but
with an increasing ratio of coarse to fine samples, with extremes of using only
fine-grained annotations and only coarse annotations on either end. We tabulate
the results of our experiments in \cref{tab:coarse_annotations_tmanet} under
the non-refined heading. We provide mean Intersection over Union (mIoU), the
standard metric for measuring segmentation performance, and mean Accuracy
(mAcc). We also show a plot of mIoU against the fraction of coarse-grained
samples used in \cref{fig:coarse_annotations_sam_miou} for better
visualisation. We also provide results for these experiments using TDNet in
\cref{tab:coarse_annotations_tdnet} and
\cref{fig:coarse_annotations_sam_miou_tdnet}.

As expected, we observe that increasing the percentage of coarse-grained
annotations causes a decrease in the model's performance. This is especially
true once there are more than 50\% coarse annotations. It is also consistent
with our earlier experiment, where we mixed 20,000 coarse and 2,975 fine
annotations. However, we note that even when we use 50\% coarse annotations, we
still get reasonable performance at a much-reduced annotation cost. According
to the Cityscapes paper~\citep{DBLP:conf/cvpr/CordtsORREBFRS16}, the time taken
for fine-grained annotation is 90 minutes per image, whereas for coarse
annotation, it is only 7 minutes.

We have established that even unrefined coarse annotations can help reduce the
overall annotation effort required. Now, we shall see how we might use SAM and
SAM 2 models to refine these coarse annotations further so that they provide
better results. As in \cref{sec:no_annotations}, we shall divide the discussion
into which approaches work and which do not.

\subsubsection{Successful Approaches}
We use the Segment Anything Model (SAM) to generate refined masks for several
segmentation classes as described in \cref{sec:method}. We tabulate the results
of our experiments with TMANet using refined coarse annotations in
\cref{tab:coarse_annotations_tmanet} under the refined heading. To ensure that
the difference in performance is not due to other factors, we use the same
samples as the non-refined setting. We similarly also plot the mIoU against the
fraction of coarse samples in \cref{fig:coarse_annotations_sam_miou}. We
observe that refined annotations perform better across all settings than coarse
annotations when using TMANet. The results are, however, not as encouraging
when we train using TDNet, shown in \cref{tab:coarse_annotations_tdnet} and
\cref{fig:coarse_annotations_sam_miou_tdnet}.. While for all coarse samples,
the refined samples still outperform the non-refined ones, at 50\% mixture, the
refined samples perform slightly worse than the non-refined ones. Due to
resource constraints, we were unable to perform more experiments with TDNet.

We also tabulate the classwise mIoU scores for a few ratios in
\cref{tab:classwise_miou} for TMANet. We observe that in most cases, the
refined coarse annotations perform better than the non-refined ones, especially
for the classes that were refined (shown in bold). Classwise scores are also
provided for TDNet in \cref{tab:classwise_miou_tdnet}. In the interest of
brevity, we only show the refined classes.

\begin{table}[t]
    \caption{Classwise Mean Intersection over Union with TDNet on the Cityscapes dataset,
        for both non-refined and refined annotations.} \centering
    \begin{tabular}{lccccc}
        \toprule
        Class    & 0\% Coarse & \multicolumn{2}{c}{50\% Coarse} & \multicolumn{2}{c}{100\% Coarse} \\
                 &            & NR    & R                       & NR    & R                        \\
        \midrule
        pole     & 51.40      & 49.31 & 50.45                   & 42.22 & 42.30                    \\
        t. light & 57.97      & 54.38 & 53.82                   & 44.44 & 46.89                    \\
        t. sign  & 69.36      & 65.64 & 67.32                   & 58.57 & 58.95                    \\
        person   & 74.36      & 72.62 & 72.16                   & 61.98 & 62.07                    \\
        rider    & 53.63      & 51.89 & 50.64                   & 40.84 & 42.68                    \\
        car      & 92.38      & 92.18 & 92.07                   & 88.57 & 88.33                    \\
        m.cycle  & 50.44      & 49.21 & 49.40                   & 38.64 & 40.26                    \\
        bicycle  & 69.41      & 67.40 & 67.23                   & 58.56 & 58.27                    \\
        \bottomrule
    \end{tabular}
    \label{tab:classwise_miou_tdnet}
\end{table}

\subsubsection{Failed Approaches}
\label{sec:coarse_annotations_what_does_not}
In the last section, we saw that refining coarse annotations for some of the
classes using SAM helps improve the performance of segmentation models when
trained with coarse samples. One inevitably asks, why limit ourselves to only
some of the classes? Why not refine masks for all classes? Unfortunately,
refining masks for classes that are not annotated as instances (such as
buildings, sky, etc.) does not work well. Also, refining classes with
inherently ambiguous boundaries, such as vegetation, does not work well. For
this reason, we only consider classes with instance-level annotations for
refinement.

We also experiment with an alternative approach for refining the coarse masks
using SAM 2. In this approach, we first use SAM 2 in automatic mask generation
mode. This provides us with a multitude of automatically generated
instance-level masks for most objects in the image. We compare each generated
mask with the coarse annotations. If more than 90\% of the labelled pixels
(pixels that have labels in the coarse annotation) in the mask belong to a
single class, we consider the entire mask to be of the same class and a
refinement of the underlying coarse annotation. The primary intuition behind
this approach was that SAM 2 would be able to generate good instance-level
masks with proper object boundaries, and then we would use the coarse
annotation to make an informed judgment as to which class the mask belongs to.
We show the results of this approach in \cref{tab:coarse_sam2}. We observe that
when we train with only coarse annotations, we get better performance when we
use these refined masks, but as we start incorporating finely-grained
annotations, our results suffer. This indicates that while the refined masks
are better than pure coarse annotations, they have enough noise, which prevents
the networks from learning good representations when we provide proper fine
annotations.

\begin{table}[t]
    \centering
    \caption{Mean IoU and Accuracy for increasing fraction of non-refined and refined coarse
        annotations with TMANet using SAM 2.}
    \begin{tabular}{Sccccc}
        \toprule
        {Coarse} \% & \multicolumn{2}{c}{Non-refined} & \multicolumn{2}{c}{Refined} \\
                    & mIoU  & mAcc                    & mIoU  & mAcc                \\
        \midrule
        0           & 76.95 & 86.19                   & 76.95 & 86.19               \\
        25          & 75.12 & 84.95                   & 66.30 & 73.59               \\
        50          & 71.13 & 79.32                   & 68.97 & 79.68               \\
        75          & 67.00 & 76.20                   & 68.70 & 78.58               \\
        100         & 54.71 & 67.18                   & 57.40 & 66.46               \\
        \bottomrule
    \end{tabular}
    \label{tab:coarse_sam2}
\end{table}

\section{Conclusion \& Future Work}
In this work, we have discussed approaches to using segmentation foundation
models, SAM and SAM 2, to reduce the effort required for manually annotating a
large number of images required for video segmentation.

We have primarily provided two ways of doing so. Either generating
pseudo-labels of unannotated future and past frames provided a manually
annotated frame in a video clip, or refining coarse annotations to help improve
performance when using them for training, mixed with fine-grained annotations.
From our experiments, we conclude that the best advice for practitioners is to
ignore the coarse annotation and only focus on generating pseudo-labels of
unannotated frames with a one-third reduction of annotation effort.

In addition to approaches that provide good results, we also discuss related
approaches that fail to do so. In performing such an analysis, we discover
certain important traits, such as diversity being important for training video
segmentation networks.

Throughout this work, our use of SAM and SAM 2 is primarily limited to treating
them as a black box for generating segmentation masks from given prompts. Thus,
a possible future direction would be to test how other segmentation models fare
in generating such pseudo-labels. Similarly, we would also like to test more
and better diversity measures than simple optical flow for selecting diverse
unannotated frames for pseudo-label generation. With better measures, our
attempts to select diverse frames may be successful rather than failures, and
thus lead to better performance.

We hope our contributions will help further research in the field of
data-efficient video segmentation and allow for low-data training.

\bibliographystyle{ACM-Reference-Format}
\bibliography{main}

@String{Computer = "{IEEE} Computer" }

@String{Springer = "Springer-Verlag" }

@inproceedings{DBLP:conf/bmvc/HungTLL018,
  author     = {Wei{-}Chih Hung and
                Yi{-}Hsuan Tsai and
                Yan{-}Ting Liou and
                Yen{-}Yu Lin and
                Ming{-}Hsuan Yang},
  title      = {Adversarial Learning for Semi-supervised Semantic Segmentation},
  _booktitle = {British Machine Vision Conference 2018, {BMVC} 2018},
  booktitle  = {{BMVC} 2018},
  _address   = {Newcastle, UK},
  address    = {UK},
  pages      = {65},
  publisher  = {{BMVA} Press},
  year       = {2018}
}

@inproceedings{DBLP:conf/cvpr/0001SRSNTC19,
  author     = {Yi Zhu and
                Karan Sapra and
                Fitsum A. Reda and
                Kevin J. Shih and
                Shawn D. Newsam and
                Andrew Tao and
                Bryan Catanzaro},
  title      = {Improving Semantic Segmentation via Video Propagation and Label
                Relaxation},
  _booktitle = {{IEEE} Conference on Computer Vision and Pattern Recognition,
                {CVPR} 2019},
  booktitle  = {{CVPR} 2019},
  _address   = {Long Beach, CA, USA},
  address    = {USA},
  pages      = {8856--8865},
  publisher  = {Computer Vision Foundation / {IEEE}},
  year       = {2019},
  _doi       = {10.1109/CVPR.2019.00906}
}

@inproceedings{DBLP:conf/cvpr/CordtsORREBFRS16,
  author     = {Marius Cordts and
                Mohamed Omran and
                Sebastian Ramos and
                Timo Rehfeld and
                Markus Enzweiler and
                Rodrigo Benenson and
                Uwe Franke and
                Stefan Roth and
                Bernt Schiele},
  title      = {The Cityscapes Dataset for Semantic Urban Scene Understanding},
  _booktitle = {2016 {IEEE} Conference on Computer Vision and Pattern
                Recognition, {CVPR} 2016},
  booktitle  = {{CVPR} 2016},
  _address   = {Las Vegas, NV, USA},
  address    = {USA},
  pages      = {3213--3223},
  publisher  = {{IEEE} Computer Society},
  year       = {2016},
  _doi       = {10.1109/CVPR.2016.350}
}

@inproceedings{DBLP:conf/cvpr/GuKW00CLCP22,
  author     = {Jiaqi Gu and
                Hyoukjun Kwon and
                Dilin Wang and
                Wei Ye and
                Meng Li and
                Yu{-}Hsin Chen and
                Liangzhen Lai and
                Vikas Chandra and
                David Z. Pan},
  title      = {Multi-Scale High-Resolution Vision Transformer for Semantic
                Segmentation},
  _booktitle = {{IEEE/CVF} Conference on Computer Vision and Pattern Recognition,
                {CVPR} 2022},
  booktitle  = {{CVPR} 2022},
  _address   = {Las Vegas, NV, USA},
  address    = {USA},
  pages      = {12084--12093},
  publisher  = {{IEEE}},
  year       = {2022},
  _doi       = {10.1109/CVPR52688.2022.01178}
}

@inproceedings{DBLP:conf/cvpr/HeZRS16,
  author     = {Kaiming He and
                Xiangyu Zhang and
                Shaoqing Ren and
                Jian Sun},
  title      = {Deep Residual Learning for Image Recognition},
  _booktitle = {2016 {IEEE} Conference on Computer Vision and Pattern
                Recognition, {CVPR} 2016},
  booktitle  = {{CVPR} 2016},
  _address   = {Las Vegas, NV, USA},
  address    = {USA},
  pages      = {770--778},
  publisher  = {{IEEE} Computer Society},
  year       = {2016},
  _doi       = {10.1109/CVPR.2016.90}
}

@inproceedings{DBLP:conf/cvpr/HuCWLSP20,
  author     = {Ping Hu and
                Fabian Caba and
                Oliver Wang and
                Zhe Lin and
                Stan Sclaroff and
                Federico Perazzi},
  title      = {Temporally Distributed Networks for Fast Video Semantic
                Segmentation},
  _booktitle = {2020 {IEEE/CVF} Conference on Computer Vision and Pattern
                Recognition, {CVPR} 2020},
  booktitle  = {{CVPR} 2020},
  _address   = {Las Vegas, NV, USA},
  address    = {USA},
  pages      = {8815--8824},
  publisher  = {Computer Vision Foundation / {IEEE}},
  year       = {2020},
  _doi       = {10.1109/CVPR42600.2020.00884}
}

@inproceedings{DBLP:conf/cvpr/Jain0C0OS23,
  author     = {Jitesh Jain and
                Jiachen Li and
                MangTik Chiu and
                Ali Hassani and
                Nikita Orlov and
                Humphrey Shi},
  title      = {OneFormer: One Transformer to Rule Universal Image Segmentation},
  _booktitle = {{IEEE/CVF} Conference on Computer Vision and Pattern Recognition,
                {CVPR} 2023},
  booktitle  = {{CVPR} 2023},
  _address   = {Vancouver, BC, Canada},
  address    = {Canada},
  pages      = {2989--2998},
  publisher  = {{IEEE}},
  year       = {2023},
  _doi       = {10.1109/CVPR52729.2023.00292}
}

@inproceedings{DBLP:conf/cvpr/JainWG19,
  author     = {Samvit Jain and
                Xin Wang and
                Joseph E. Gonzalez},
  title      = {Accel: {A} Corrective Fusion Network for Efficient Semantic
                Segmentation on Video},
  _booktitle = {{IEEE} Conference on Computer Vision and Pattern Recognition,
                {CVPR} 2019},
  booktitle  = {{CVPR} 2019},
  _address   = {Long Beach, CA, USA},
  address    = {USA},
  pages      = {8866--8875},
  publisher  = {Computer Vision Foundation / {IEEE}},
  year       = {2019},
  _doi       = {10.1109/CVPR.2019.00907}
}

@inproceedings{DBLP:conf/cvpr/LiSL18,
  author     = {Yule Li and
                Jianping Shi and
                Dahua Lin},
  title      = {Low-Latency Video Semantic Segmentation},
  _booktitle = {2018 {IEEE} Conference on Computer Vision and Pattern
                Recognition, {CVPR} 2018},
  booktitle  = {{CVPR} 2018},
  _address   = {Salt Lake City, UT, USA},
  address    = {USA},
  pages      = {5997--6005},
  publisher  = {Computer Vision Foundation / {IEEE} Computer Society},
  year       = {2018},
  _doi       = {10.1109/CVPR.2018.00628}
}

@inproceedings{DBLP:conf/cvpr/LiuWQYBS17,
  author     = {Si Liu and
                Changhu Wang and
                Ruihe Qian and
                Han Yu and
                Renda Bao and
                Yao Sun},
  title      = {Surveillance Video Parsing with Single Frame Supervision},
  _booktitle = {2017 {IEEE} Conference on Computer Vision and Pattern
                Recognition, {CVPR} 2017},
  booktitle  = {{CVPR} 2017},
  _address   = {Honolulu, HI, USA},
  address    = {USA},
  pages      = {1013--1021},
  publisher  = {{IEEE} Computer Society},
  year       = {2017},
  _doi       = {10.1109/CVPR.2017.114}
}

@inproceedings{DBLP:conf/cvpr/LongSD15,
  author     = {Jonathan Long and
                Evan Shelhamer and
                Trevor Darrell},
  title      = {Fully convolutional networks for semantic segmentation},
  _booktitle = {{IEEE} Conference on Computer Vision and Pattern Recognition,
                {CVPR} 2015},
  booktitle  = {{CVPR} 2015},
  _address   = {Boston, MA, USA},
  address    = {USA},
  pages      = {3431--3440},
  publisher  = {{IEEE} Computer Society},
  year       = {2015},
  _doi       = {10.1109/CVPR.2015.7298965}
}

@inproceedings{DBLP:conf/cvpr/NilssonS18,
  author     = {David Nilsson and
                Cristian Sminchisescu},
  title      = {Semantic Video Segmentation by Gated Recurrent Flow Propagation},
  _booktitle = {2018 {IEEE} Conference on Computer Vision and Pattern
                Recognition, {CVPR} 2018},
  booktitle  = {{CVPR} 2018},
  _address   = {Salt Lake City, UT, USA},
  address    = {USA},
  pages      = {6819--6828},
  publisher  = {Computer Vision Foundation / {IEEE} Computer Society},
  year       = {2018},
  _doi       = {10.1109/CVPR.2018.00713}
}

@inproceedings{DBLP:conf/cvpr/SunLDPG22,
  author     = {Guolei Sun and
                Yun Liu and
                Henghui Ding and
                Thomas Probst and
                Luc Van Gool},
  title      = {Coarse-to-Fine Feature Mining for Video Semantic Segmentation},
  _booktitle = {{IEEE/CVF} Conference on Computer Vision and Pattern Recognition,
                {CVPR} 2022},
  booktitle  = {{CVPR} 2022},
  _address   = {Las Vegas, NV, USA},
  address    = {USA},
  pages      = {3116--3127},
  publisher  = {{IEEE}},
  year       = {2022},
  _doi       = {10.1109/CVPR52688.2022.00313}
}

@inproceedings{DBLP:conf/cvpr/WangWSFLJWZL22,
  author     = {Yuchao Wang and
                Haochen Wang and
                Yujun Shen and
                Jingjing Fei and
                Wei Li and
                Guoqiang Jin and
                Liwei Wu and
                Rui Zhao and
                Xinyi Le},
  title      = {Semi-Supervised Semantic Segmentation Using Unreliable
                Pseudo-Labels},
  _booktitle = {{IEEE/CVF} Conference on Computer Vision and Pattern Recognition,
                {CVPR} 2022},
  booktitle  = {{CVPR} 2022},
  _address   = {Las Vegas, NV, USA},
  address    = {USA},
  pages      = {4238--4247},
  publisher  = {{IEEE}},
  year       = {2022},
  _doi       = {10.1109/CVPR52688.2022.00421}
}

@inproceedings{DBLP:conf/cvpr/XieGDTH17,
  author     = {Saining Xie and
                Ross B. Girshick and
                Piotr Doll{\'{a}}r and
                Zhuowen Tu and
                Kaiming He},
  title      = {Aggregated Residual Transformations for Deep Neural Networks},
  _booktitle = {2017 {IEEE} Conference on Computer Vision and Pattern
                Recognition, {CVPR} 2017},
  booktitle  = {{CVPR} 2017},
  _address   = {Honolulu, HI, USA},
  address    = {USA},
  pages      = {5987--5995},
  publisher  = {{IEEE} Computer Society},
  year       = {2017},
  _doi       = {10.1109/CVPR.2017.634}
}

@inproceedings{DBLP:conf/cvpr/YangYZLY18,
  author     = {Maoke Yang and
                Kun Yu and
                Chi Zhang and
                Zhiwei Li and
                Kuiyuan Yang},
  title      = {DenseASPP for Semantic Segmentation in Street Scenes},
  _booktitle = {2018 {IEEE} Conference on Computer Vision and Pattern
                Recognition, {CVPR} 2018},
  booktitle  = {{CVPR} 2018},
  _address   = {Salt Lake City, UT, USA},
  address    = {USA},
  pages      = {3684--3692},
  publisher  = {Computer Vision Foundation / {IEEE} Computer Society},
  year       = {2018},
  _doi       = {10.1109/CVPR.2018.00388}
}

@inproceedings{DBLP:conf/cvpr/YuCWXCLMD20,
  author     = {Fisher Yu and
                Haofeng Chen and
                Xin Wang and
                Wenqi Xian and
                Yingying Chen and
                Fangchen Liu and
                Vashisht Madhavan and
                Trevor Darrell},
  title      = {{BDD100K:} {A} Diverse Driving Dataset for Heterogeneous
                Multitask Learning},
  _booktitle = {2020 {IEEE/CVF} Conference on Computer Vision and Pattern
                Recognition, {CVPR} 2020},
  booktitle  = {{CVPR} 2020},
  _address   = {Las Vegas, NV, USA},
  address    = {USA},
  pages      = {2633--2642},
  publisher  = {Computer Vision Foundation / {IEEE}},
  year       = {2020},
  _doi       = {10.1109/CVPR42600.2020.00271}
}

@inproceedings{DBLP:conf/cvpr/YuKF17,
  author     = {Fisher Yu and
                Vladlen Koltun and
                Thomas A. Funkhouser},
  title      = {Dilated Residual Networks},
  _booktitle = {2017 {IEEE} Conference on Computer Vision and Pattern
                Recognition, {CVPR} 2017},
  booktitle  = {{CVPR} 2017},
  _address   = {Honolulu, HI, USA},
  address    = {USA},
  pages      = {636--644},
  publisher  = {{IEEE} Computer Society},
  year       = {2017},
  _doi       = {10.1109/CVPR.2017.75}
}

@inproceedings{DBLP:conf/cvpr/ZhaoSQWJ17,
  author     = {Hengshuang Zhao and
                Jianping Shi and
                Xiaojuan Qi and
                Xiaogang Wang and
                Jiaya Jia},
  title      = {Pyramid Scene Parsing Network},
  _booktitle = {2017 {IEEE} Conference on Computer Vision and Pattern
                Recognition, {CVPR} 2017},
  booktitle  = {{CVPR} 2017},
  _address   = {Honolulu, HI, USA},
  address    = {USA},
  pages      = {6230--6239},
  publisher  = {{IEEE} Computer Society},
  year       = {2017},
  _doi       = {10.1109/CVPR.2017.660}
}

@inproceedings{DBLP:conf/eccv/BrostowSFC08,
  author     = {Gabriel J. Brostow and
                Jamie Shotton and
                Julien Fauqueur and
                Roberto Cipolla},
  editor     = {David A. Forsyth and
                Philip H. S. Torr and
                Andrew Zisserman},
  title      = {Segmentation and Recognition Using Structure from Motion Point
                Clouds},
  _booktitle = {Computer Vision - {ECCV} 2008, 10th European Conference on
                Computer Vision, Proceedings, Part {I}},
  booktitle  = {{ECCV} 2008},
  _address   = {Marseille, France},
  address    = {France},
  series     = {Lecture Notes in Computer Science},
  volume     = {5302},
  pages      = {44--57},
  publisher  = {Springer},
  year       = {2008},
  _doi       = {10.1007/978-3-540-88682-2\_5}
}

@inproceedings{DBLP:conf/eccv/ChenLCCCZAS20,
  author     = {Liang{-}Chieh Chen and
                Raphael Gontijo Lopes and
                Bowen Cheng and
                Maxwell D. Collins and
                Ekin D. Cubuk and
                Barret Zoph and
                Hartwig Adam and
                Jonathon Shlens},
  title      = {Naive-Student: Leveraging Semi-Supervised Learning in Video
                Sequences for Urban Scene Segmentation},
  _booktitle = {Computer Vision - {ECCV} 2020 - 16th European Conference,
                Proceedings, Part {IX}},
  booktitle  = {{ECCV} 2020},
  _address   = {Glasgow, UK},
  address    = {UK},
  series     = {Lecture Notes in Computer Science},
  volume     = {12354},
  pages      = {695--714},
  publisher  = {Springer},
  year       = {2020},
  _doi       = {10.1007/978-3-030-58545-7\_40}
}

@inproceedings{DBLP:conf/eccv/LiuSYW20,
  author     = {Yifan Liu and
                Chunhua Shen and
                Changqian Yu and
                Jingdong Wang},
  editor     = {Andrea Vedaldi and
                Horst Bischof and
                Thomas Brox and
                Jan{-}Michael Frahm},
  title      = {Efficient Semantic Video Segmentation with Per-Frame Inference},
  _booktitle = {Computer Vision - {ECCV} 2020 - 16th European Conference,
                Proceedings, Part {X}},
  booktitle  = {{ECCV} 2020},
  _address   = {Glasgow, UK},
  address    = {UK},
  series     = {Lecture Notes in Computer Science},
  volume     = {12355},
  pages      = {352--368},
  publisher  = {Springer},
  year       = {2020},
  _doi       = {10.1007/978-3-030-58607-2\_21}
}

@inproceedings{DBLP:conf/eccv/LuoY20,
  author     = {Wenfeng Luo and
                Meng Yang},
  editor     = {Andrea Vedaldi and
                Horst Bischof and
                Thomas Brox and
                Jan{-}Michael Frahm},
  title      = {Semi-supervised Semantic Segmentation via Strong-Weak Dual-Branch
                Network},
  _booktitle = {Computer Vision - {ECCV} 2020 - 16th European Conference,
                Proceedings, Part {V}},
  booktitle  = {{ECCV} 2020},
  _address   = {Glasgow, UK},
  address    = {UK},
  series     = {Lecture Notes in Computer Science},
  volume     = {12350},
  pages      = {784--800},
  publisher  = {Springer},
  year       = {2020},
  _doi       = {10.1007/978-3-030-58558-7\_46}
}

@inproceedings{DBLP:conf/iccv/GaddeJG17,
  author     = {Raghudeep Gadde and
                Varun Jampani and
                Peter V. Gehler},
  title      = {Semantic Video CNNs Through Representation Warping},
  _booktitle = {{IEEE} International Conference on Computer Vision, {ICCV} 2017},
  booktitle  = {{ICCV} 2017},
  _address   = {Venice, Italy},
  address    = {Italy},
  pages      = {4463--4472},
  publisher  = {{IEEE} Computer Society},
  year       = {2017},
  _doi       = {10.1109/ICCV.2017.477}
}

@inproceedings{DBLP:conf/iccv/KirillovMRMRGXW23,
  author     = {Alexander Kirillov and
                Eric Mintun and
                Nikhila Ravi and
                Hanzi Mao and
                Chlo{\'{e}} Rolland and
                Laura Gustafson and
                Tete Xiao and
                Spencer Whitehead and
                Alexander C. Berg and
                Wan{-}Yen Lo and
                Piotr Doll{\'{a}}r and
                Ross B. Girshick},
  title      = {Segment Anything},
  _booktitle = {{IEEE/CVF} International Conference on Computer Vision,
                {ICCV} 2023},
  booktitle  = {{ICCV} 2023},
  _address   = {Paris, France},
  address    = {France},
  pages      = {3992--4003},
  publisher  = {{IEEE}},
  year       = {2023},
  _doi       = {10.1109/ICCV51070.2023.00371}
}

@inproceedings{DBLP:conf/iccv/SoulySS17,
  author     = {Nasim Souly and
                Concetto Spampinato and
                Mubarak Shah},
  title      = {Semi Supervised Semantic Segmentation Using Generative
                Adversarial Network},
  _booktitle = {{IEEE} International Conference on Computer Vision, {ICCV} 2017},
  booktitle  = {{ICCV} 2017},
  _address   = {Venice, Italy},
  address    = {Italy},
  pages      = {5689--5697},
  publisher  = {{IEEE} Computer Society},
  year       = {2017},
  _doi       = {10.1109/ICCV.2017.606}
}

@inproceedings{DBLP:conf/iccv/StrudelPLS21,
  author     = {Robin Strudel and
                Ricardo Garcia and
                Ivan Laptev and
                Cordelia Schmid},
  title      = {Segmenter: Transformer for Semantic Segmentation},
  _booktitle = {2021 {IEEE/CVF} International Conference on Computer Vision,
                {ICCV} 2021},
  booktitle  = {{ICCV} 2021},
  _address   = {Montreal, QC, Canada},
  address    = {Canada},
  pages      = {7242--7252},
  publisher  = {{IEEE}},
  year       = {2021},
  _doi       = {10.1109/ICCV48922.2021.00717}
}

@inproceedings{DBLP:conf/iccvw/JainSOHL0S23,
  author     = {Jitesh Jain and
                Anukriti Singh and
                Nikita Orlov and
                Zilong Huang and
                Jiachen Li and
                Steven Walton and
                Humphrey Shi},
  title      = {SeMask: Semantically Masked Transformers for Semantic
                Segmentation},
  _booktitle = {{IEEE/CVF} International Conference on Computer Vision, {ICCV}
                2023 - Workshops},
  booktitle  = {{ICCV} 2023 - Workshops},
  _address   = {Paris, France},
  address    = {France},
  pages      = {752--761},
  publisher  = {{IEEE}},
  year       = {2023},
  _doi       = {10.1109/ICCVW60793.2023.00083}
}

@inproceedings{DBLP:conf/icip/WangW021,
  author     = {Hao Wang and
                Weining Wang and
                Jing Liu},
  title      = {Temporal Memory Attention for Video Semantic Segmentation},
  _booktitle = {2021 {IEEE} International Conference on Image Processing, {ICIP}
                2021},
  booktitle  = {{ICIP} 2021},
  _address   = {Anchorage, AK, USA},
  address    = {USA},
  pages      = {2254--2258},
  publisher  = {{IEEE}},
  year       = {2021},
  _doi       = {10.1109/ICIP42928.2021.9506731}
}

@inproceedings{DBLP:conf/iclr/DosovitskiyB0WZ21,
  author     = {Alexey Dosovitskiy and
                Lucas Beyer and
                Alexander Kolesnikov and
                Dirk Weissenborn and
                Xiaohua Zhai and
                Thomas Unterthiner and
                Mostafa Dehghani and
                Matthias Minderer and
                Georg Heigold and
                Sylvain Gelly and
                Jakob Uszkoreit and
                Neil Houlsby},
  title      = {An Image is Worth 16x16 Words: Transformers for Image Recognition
                at Scale},
  _booktitle = {9th International Conference on Learning Representations, {ICLR}
                2021},
  booktitle  = {{ICLR} 2021},
  _address   = {Virtual Event, Austria},
  address    = {Virtual Event},
  publisher  = {OpenReview.net},
  year       = {2021}
}

@inproceedings{DBLP:conf/iclr/RaviGHHR0KRRGMP25,
  author     = {Nikhila Ravi and
                Valentin Gabeur and
                Yuan{-}Ting Hu and
                Ronghang Hu and
                Chaitanya Ryali and
                Tengyu Ma and
                Haitham Khedr and
                Roman R{\"{a}}dle and
                Chlo{\'{e}} Rolland and
                Laura Gustafson and
                Eric Mintun and
                Junting Pan and
                Kalyan Vasudev Alwala and
                Nicolas Carion and
                Chao{-}Yuan Wu and
                Ross B. Girshick and
                Piotr Doll{\'{a}}r and
                Christoph Feichtenhofer},
  title      = {{SAM} 2: Segment Anything in Images and Videos},
  _booktitle = {The Thirteenth International Conference on Learning
                Representations, {ICLR} 2025},
  booktitle  = {{ICLR} 2025},
  address    = {Singapore},
  publisher  = {OpenReview.net},
  year       = {2025}
}

@inproceedings{DBLP:conf/iclr/ZouZZLBHP21,
  author     = {Yuliang Zou and
                Zizhao Zhang and
                Han Zhang and
                Chun{-}Liang Li and
                Xiao Bian and
                Jia{-}Bin Huang and
                Tomas Pfister},
  title      = {PseudoSeg: Designing Pseudo Labels for Semantic Segmentation},
  _booktitle = {9th International Conference on Learning Representations, {ICLR}
                2021},
  booktitle  = {{ICLR} 2021},
  _address   = {Virtual Event, Austria},
  address    = {Virtual Event},
  publisher  = {OpenReview.net},
  year       = {2021}
}

@inproceedings{DBLP:conf/icmcs/LeeCP21,
  author     = {Shih{-}Po Lee and
                Si{-}Cun Chen and
                Wen{-}Hsiao Peng},
  title      = {{GSVNET:} Guided Spatially-Varying Convolution for Fast Semantic
                Segmentation on Video},
  _booktitle = {2021 {IEEE} International Conference on Multimedia and Expo,
                {ICME} 2021},
  booktitle  = {{ICME} 2021},
  _address   = {Shenzhen, China},
  address    = {China},
  pages      = {1--6},
  publisher  = {{IEEE}},
  year       = {2021},
  _doi       = {10.1109/ICME51207.2021.9428381}
}

@inproceedings{DBLP:conf/icml/RadfordKHRGASAM21,
  author     = {Alec Radford and
                Jong Wook Kim and
                Chris Hallacy and
                Aditya Ramesh and
                Gabriel Goh and
                Sandhini Agarwal and
                Girish Sastry and
                Amanda Askell and
                Pamela Mishkin and
                Jack Clark and
                Gretchen Krueger and
                Ilya Sutskever},
  title      = {Learning Transferable Visual Models From Natural Language
                Supervision},
  _booktitle = {Proceedings of the 38th International Conference on Machine
                Learning, {ICML} 2021},
  booktitle  = {{ICML} 2021},
  address    = {Virtual Event},
  series     = {Proceedings of Machine Learning Research},
  volume     = {139},
  pages      = {8748--8763},
  publisher  = {{PMLR}},
  year       = {2021}
}

@inproceedings{DBLP:conf/nips/ChengSK21,
  author     = {Bowen Cheng and
                Alexander G. Schwing and
                Alexander Kirillov},
  title      = {Per-Pixel Classification is Not All You Need for Semantic
                Segmentation},
  _booktitle = {Advances in Neural Information Processing Systems 34: Annual
                Conference on Neural Information Processing Systems 2021, NeurIPS
                2021},
  booktitle  = {NeurIPS 2021},
  address    = {virtual},
  pages      = {17864--17875},
  year       = {2021}
}

@inproceedings{DBLP:conf/nips/VaswaniSPUJGKP17,
  author     = {Ashish Vaswani and
                Noam Shazeer and
                Niki Parmar and
                Jakob Uszkoreit and
                Llion Jones and
                Aidan N. Gomez and
                Lukasz Kaiser and
                Illia Polosukhin},
  title      = {Attention is All you Need},
  _booktitle = {Advances in Neural Information Processing Systems 30: Annual
                Conference on Neural Information Processing Systems 2017},
  booktitle  = {NeurIPS 2017},
  _address   = {Long Beach, CA, {USA}},
  address    = {{USA}},
  pages      = {5998--6008},
  year       = {2017}
}

@inproceedings{DBLP:conf/nips/XieWYAAL21,
  author     = {Enze Xie and
                Wenhai Wang and
                Zhiding Yu and
                Anima Anandkumar and
                Jos{\'{e}} M. {\'{A}}lvarez and
                Ping Luo},
  title      = {SegFormer: Simple and Efficient Design for Semantic Segmentation
                with Transformers},
  _booktitle = {Advances in Neural Information Processing Systems 34: Annual
                Conference on Neural Information Processing Systems 2021, NeurIPS
                2021},
  booktitle  = {NeurIPS 2021},
  address    = {virtual},
  pages      = {12077--12090},
  year       = {2021}
}

@inproceedings{DBLP:conf/wacv/DasXHAS23,
  author     = {Anurag Das and
                Yongqin Xian and
                Yang He and
                Zeynep Akata and
                Bernt Schiele},
  title      = {Urban Scene Semantic Segmentation with Low-Cost Coarse
                Annotation},
  _booktitle = {{IEEE/CVF} Winter Conference on Applications of Computer Vision,
                {WACV} 2023},
  booktitle  = {{WACV} 2023},
  _address   = {Waikoloa, HI, USA},
  address    = {USA},
  pages      = {5967--5976},
  publisher  = {{IEEE}},
  year       = {2023},
  _doi       = {10.1109/WACV56688.2023.00592}
}

@inproceedings{DBLP:conf/wacv/VarmaSNCJ19,
  author     = {Girish Varma and
                Anbumani Subramanian and
                Anoop M. Namboodiri and
                Manmohan Chandraker and
                C. V. Jawahar},
  title      = {{IDD:} {A} Dataset for Exploring Problems of Autonomous
                Navigation in Unconstrained Environments},
  _booktitle = {{IEEE} Winter Conference on Applications of Computer Vision,
                {WACV} 2019},
  booktitle  = {{WACV} 2019},
  _address   = {Waikoloa Village, HI, USA},
  address    = {USA},
  pages      = {1743--1751},
  publisher  = {{IEEE}},
  year       = {2019},
  _doi       = {10.1109/WACV.2019.00190}
}

@article{DBLP:journals/corr/abs-2004-14960,
  author     = {Yi Zhu and
                Zhongyue Zhang and
                Chongruo Wu and
                Zhi Zhang and
                Tong He and
                Hang Zhang and
                R. Manmatha and
                Mu Li and
                Alexander J. Smola},
  title      = {Improving Semantic Segmentation via Self-Training},
  journal    = {CoRR},
  volume     = {abs/2004.14960},
  year       = {2020},
  url        = {https://arxiv.org/abs/2004.14960},
  eprinttype = {arXiv},
  eprint     = {2004.14960}
}

@inproceedings{DBLP:journals/corr/ChenPKMY14,
  author     = {Liang{-}Chieh Chen and
                George Papandreou and
                Iasonas Kokkinos and
                Kevin Murphy and
                Alan L. Yuille},
  title      = {Semantic Image Segmentation with Deep Convolutional Nets and
                Fully Connected CRFs},
  _booktitle = {3rd International Conference on Learning Representations, {ICLR}
                2015, Conference Track Proceedings},
  booktitle  = {{ICLR} 2015},
  _address   = {San Diego, CA, USA},
  address    = {USA},
  year       = {2015}
}

@inproceedings{DBLP:journals/corr/SimonyanZ14a,
  author     = {Karen Simonyan and
                Andrew Zisserman},
  title      = {Very Deep Convolutional Networks for Large-Scale Image
                Recognition},
  _booktitle = {3rd International Conference on Learning Representations, {ICLR}
                2015, Conference Track Proceedings},
  booktitle  = {{ICLR} 2015},
  _address   = {San Diego, CA, USA},
  address    = {USA},
  year       = {2015}
}

@article{DBLP:journals/ijmir/GuoLGL18,
  author  = {Yanming Guo and
             Yu Liu and
             Theodoros Georgiou and
             Michael S. Lew},
  title   = {A review of semantic segmentation using deep neural networks},
  journal = {Int. J. Multim. Inf. Retr.},
  volume  = {7},
  number  = {2},
  pages   = {87--93},
  year    = {2018},
  _doi    = {10.1007/S13735-017-0141-Z}
}

@article{DBLP:journals/ijon/HaoZG20,
  author  = {Shijie Hao and
             Yuan Zhou and
             Yanrong Guo},
  title   = {A Brief Survey on Semantic Segmentation with Deep Learning},
  journal = {Neurocomputing},
  volume  = {406},
  pages   = {302--321},
  year    = {2020},
  _doi    = {10.1016/J.NEUCOM.2019.11.118}
}

@article{DBLP:journals/pami/ChenPKMY18,
  author  = {Liang{-}Chieh Chen and
             George Papandreou and
             Iasonas Kokkinos and
             Kevin Murphy and
             Alan L. Yuille},
  title   = {DeepLab: Semantic Image Segmentation with Deep Convolutional Nets,
             Atrous Convolution, and Fully Connected CRFs},
  journal = {{IEEE} Trans. Pattern Anal. Mach. Intell.},
  volume  = {40},
  number  = {4},
  pages   = {834--848},
  year    = {2018},
  _doi    = {10.1109/TPAMI.2017.2699184}
}

@article{DBLP:journals/pami/MittalTB21,
  author  = {Sudhanshu Mittal and
             Maxim Tatarchenko and
             Thomas Brox},
  title   = {Semi-Supervised Semantic Segmentation With High- and Low-Level
             Consistency},
  journal = {{IEEE} Trans. Pattern Anal. Mach. Intell.},
  volume  = {43},
  number  = {4},
  pages   = {1369--1379},
  year    = {2021},
  _doi    = {10.1109/TPAMI.2019.2960224}
}

\end{document}